\documentclass[letterpaper, 10 pt, conference]{ieeeconf}  % Comment this line out if you need a4paper

\usepackage{color}
\IEEEoverridecommandlockouts                              % This command is only needed if 
\usepackage{graphics} % for pdf, bitmapped graphics files
\usepackage{epsfig} % for postscript graphics files
\usepackage{amsmath} % assumes amsmath package installed
\usepackage{amssymb}  % assumes amsmath package installed
\usepackage{amsbsy}
\usepackage{mathtools}
\usepackage{bm}
\usepackage{subfiles}
\usepackage[noend,ruled,linesnumbered]{algorithm2e}
\usepackage{caption}
\usepackage{subcaption}
\usepackage{cite}

\DeclareCaptionFont{mysize}{\fontsize{8}{9.6}\selectfont}
\newtheorem{assumption}{\textbf{Assumption}}

\newtheorem{lemma}{\textbf{Lemma}}
\newtheorem{remark}{\textbf{Remark}}

\newtheorem{theorem}{\textbf{Theorem}}
\newtheorem{cor}{\textbf{Corollary}}

\newcommand{\mbb}[1]{\mathbb{#1}}
\newcommand{\mcal}[1]{\mathcal{#1}}
\newcommand{\mbf}[1]{\mathbf{#1}}
\newcommand{\qnew}{\mbf{q}_{i, \text{new}}}
\newcommand{\Sigmanew}{\Sigma_{i, \text{new}}}
\newcommand{\pnew}{\mbf{p}_{i, \text{new}}}
\newcommand{\qrand}{\mbf{q}_{i, \text{rand}}}
\newcommand{\Sigmarand}{\Sigma_{i, \text{rand}}}
\newcommand{\prand}{\mbf{p}_{i, \text{rand}}}
\newcommand{\unew}{\mbf{u}_{i, \text{new}}}

\graphicspath{figures}

\title{\LARGE \bf
Technical Report: Distributed Sampling-based Planning for \\Non-Myopic Active Information Gathering
}

\author{Mariliza Tzes$^{1}$, Yiannis Kantaros$^{1}$, George J. Pappas$^{1}$% <-this % stops a space
\thanks{This work was supported by the ARL grant DCIST CRA W911NF-17-2-0181.
}% <-this % stops a space
\thanks{$^{1}$M. Tzes, Y. Kantaros and G. Pappas are with GRASP Lab, University of Pennsylvania, Philadelphia, PA 19104, USA, {\tt\small \{mtzes, kantaros, pappasg\}@seas.upenn.edu}.}
}

\begin{document}

\maketitle
\thispagestyle{empty}
\pagestyle{empty}

%%%%%%%%%%%%%%%%%%%%%%%%%%%%%%%%%%%%%%%%%%%%%%%%%%%%%%%%%%%%%%%%%%%%%%%%%%%%%%%%
\begin{abstract}
%\textcolor{red}{We can edit the abstract later once the paper is finalized}
This paper addresses the problem of active information gathering for multi-robot systems. Specifically, we consider scenarios where robots are tasked with reducing uncertainty of dynamical hidden states evolving in complex environments. The majority of existing information gathering approaches are centralized and, therefore, they cannot be applied to distributed robot teams where communication to a central user is not available. To address this challenge, we propose a novel distributed sampling-based planning algorithm that can significantly increase robot and target scalability while decreasing computational cost. In our non-myopic approach, all robots build in parallel local trees exploring the information space and their corresponding motion space. As the robots construct their respective local trees, they communicate with their neighbors to exchange and aggregate their local beliefs about the hidden state through a distributed Kalman filter. We show that the proposed algorithm is probabilistically complete and asymptotically optimal. We provide extensive simulation results that demonstrate the scalability of the proposed algorithm and that it can address large-scale, multi-robot information gathering tasks, that are computationally challenging for centralized methods.
%\textcolor{red}{Yiannis will use red color for comments/notes}\textcolor{blue}{Mariliza will use blue color for comments/notes}
\end{abstract}

%%%%%%%%%%%%%%%%%%%%%%%%%%%%%%%%%%%%%%%%%%%%%%%%%%%%%%%%%%%%%%%%%%%%%%%%%%%%%%%%

%\subfile{introduction.tex}
\section{INTRODUCTION}

%This template provides authors with most of the formatting specifications needed for preparing electronic versions of their papers. All standard paper components have been specified for three reasons: (1) ease of use when formatting individual papers, (2) automatic compliance to electronic requirements that facilitate the concurrent or later production of electronic products, and (3) conformity of style throughout a conference proceedings. Margins, column widths, line spacing, and type styles are built-in; examples of the type styles are provided throughout this document and are identified in italic type, within parentheses, following the example. Some components, such as multi-leveled equations, graphics, and tables are not prescribed, although the various table text styles are provided. The formatter will need to create these components, incorporating the applicable criteria that follow.

%1) Introductory paragraph
Recent advances in sensing and perception have enabled the deployment of mobile robots in unknown environments for information gathering missions such as environmental monitoring \cite{ma2017informative}, \cite{lu2018mobile}, surveillance \cite{frew2008target}, coverage \cite{tzes2018visual}, \cite{kusnur2020search}, target tracking \cite{bucci2019decentralized},\cite{huang2015bank} and active-SLAM \cite{carlone2014active}, \cite{atanasov2015decentralized}. These tasks require spatio-temporal information collection, which can be achieved more efficiently by multi-robot systems, rather than relying on individual robots. To avoid the need for a central user computing sensor-based control policies for large multi-robot systems, distributed control frameworks are needed allowing robots to make decisions locally.  
%Multi-agent systems have been deployed for various  cooperative  information  gathering  missions such as environmental monitoring \cite{ma2017informative}, \cite{lu2018mobile}, surveillance \cite{frew2008target}, coverage \cite{tzes2018visual}, \cite{kusnur2020search}, target tracking \cite{bucci2019decentralized}, \cite{huang2015bank} and active-SLAM \cite{carlone2014active}, \cite{atanasov2015decentralized}. Such applications appertain to the wider problem of Active Information Acquisition (AIA) where robots collaboratively collect information about a physical phenomenon of interest.

%2) In this paper... (what problem we solve and how)
In this paper, we are interested in designing \textit{distributed} control policies for a team of mobile sensing robots tasked with actively reducing the accumulated uncertainty of a dynamic hidden state over an a priori unknown horizon while satisfying user-specified accuracy requirements.  
In particular, first we formulate the Active Information Acquisition (AIA) problem as a stochastic optimal control problem. Then, building upon the separation principle presented in \cite{atanasov2014information}, we convert the stochastic optimal control problem into a deterministic optimal control problem via the use of a Distributed Kalman Filter (DKF) for which offline/open-loop control policies are optimal.
To solve the resulting deterministic optimal control problem, we propose a novel distributed sampling-based method under which all robot incrementally and in parallel build their own directed trees that explore both their respective robot motion and information space. To build these trees, the robots communicate with each other over an underlying connected communication network to exchange their beliefs about the hidden state which are then fused using a DKF \cite{atanasov2014joint}.
%Robots exchange certain information with their neighbors and update their uncertainty according to a distributed bayesian filter which under linear, gaussian assumptions coincides with the Distributed Kalman Filter (DKF) \cite{atanasov2014joint}. We formulate the AIA as a stochastic optimal control problem and we prove that under the aforementioned assumptions the problem is converted into a deterministic optimal control problem by extending the separation principle \cite{atanasov2014information} to work along with DKF. 
We show that the proposed algorithm is probabilistically complete and asymptotically optimal. The proposed scheme is evaluated through extensive simulations for a target localization and tracking scenario.

{\bf{Literature Review:}} Relevant works  that  address informative planning problems can be categorized into myopic/greedy and non-myopic (sampling-based and search-based). Myopic approaches typically rely on gradient-based controllers that although they enjoy computational efficiency, they suffer from local optima \cite{chung2006decentralized, olfati2007distributed, hoffmann2009mobile, dames2012decentralized, charrow2014approximate, meyer2015distributed, corah2018distributed}. In \cite{dames2012decentralized} a \textit{decentralized}, myopic approach is introduced where the robots are driven along the gradient of the Mutual Information (MI) between the targets and the sensor observations.
To mitigate the issue of local optimality, non-myopic \textit{search-based} approaches have been proposed that can design optimal paths  \cite{le2009trajectory}. Typically, these methods are computationally expensive as they require exploring exhaustively both the robot motion and the information space in a centralized fashion. More computationally efficient but suboptimal controllers have also been proposed that rely on pruning the exploration process and on addressing the information gathering problem in a \textit{decentralized} way via coordinate descent \cite{singh2009efficient,atanasov2015decentralized,schlotfeldt2018anytime}. However, these approaches become computationally intractable as the planning horizon or the number of robots increases as decisions are made locally but \textit{sequentially} across the robots. Monte Carlo Tree Search \cite{browne2012survey} has recently gain popularity for online planning in robotics. Best et. al \cite{best2019dec} suggest the Dec-MCTS algorithm that efficiently samples the individual action space of each robot on contrary to our proposed method and then coordinates with a sparse approximation of the joint action space. To do so, the robots must exchange branches of their Monte-Carlo trees, burdening further the communication needs.
Nonmyopic \textit{sampling-based} approaches have also been proposed due to their ability to find feasible solutions very fast, see e.g., \cite{levine2010information,hollinger2014sampling,khodayi2019distributed,lan2016rapidly}. Common in these works is that they are \textit{centralized} and, therefore, as the number of robots or the dimensions of the hidden states increase, the state-space that needs to be explored grows exponentially and, as result, sampling-based approaches also fail to compute sensor policies because of either excessive runtime or memory requirements.  A more scalable but centralized sampling-based approach is proposed in \cite{kantaros2019asymptotically} that requires communication to a central user for both estimation and control purposes. Building upon our previous work \cite{kantaros2019asymptotically}, we propose a novel \textit{distributed sampling-based} algorithm that scales well with the number of robots and planning horizon while allowing the robots to \textit{locally} and \textit{simultaneously} explore their physical and reachable information space, exchange their beliefs over an underlying communication network, and fuse them via a DKF. %\textcolor{blue}{In contrast to the centralized approach in \cite{kantaros2019asymptotically}, the proposed scheme is robust to different computational capabilities among the robots.} 

{\bf{Contributions:}} The contribution of this paper can be summarized as follows.
\textit{First}, we propose a \textit{distributed} nonmyopic  sampling-based approach for information-gathering tasks. \textit{Second}, we propose the first distributed sampling-based information gathering algorithm that is probabilistically complete and asymptotically optimal, %the proposed algorithm preserves the completeness and optimality properties of its centralized counterpart \cite{kantaros2019asymptotically}
while it signfinicantly decreases the computational complexity per iteration of its centralized counterpart \cite{kantaros2019asymptotically}.
%\textcolor{blue}{\textit{Third} we prove that the proposed scheme decreases the computational complexity per iteration of its centralized counterpart \cite{kantaros2019asymptotically}.}
\textit{Third},  we provide extensive simulation results that show that the proposed method can efficiently handle large-scale estimation tasks.
%Our previous work \cite{kantaros2019asymptotically} developed a sampling-based AIA algorithm which is centralized in control and estimation and assumes that robots have access to a central source of information. In this paper we propose a novel distributed sampling-based AIA scheme that
%\begin{itemize}
%    \item is highly scalable with the increase of number of robots, dimension of the hiddens state
%    \item is completely parallelizable
%    \item incorporates a distributed information filter to relax
%the previous requirement of each robot having access to
%centralized information 
%    \item is proven to be asymptotically complete and optimal
%\end{itemize}

\section{Problem Formulation}
\subsection{Robot Dynamics, Hidden State, and Observation Model}
%%%%%% environment %%%%%%%
Consider a team of $N$ mobile robots that reside in a complex environment $\Omega \subset \mathbb{R}^d$ where $d$ is the dimension of the workspace. Obstacles of arbitrary shape are denoted as $\mathcal{O}$ and the obstacle-free area is $\Omega_{\text{free}}\coloneqq \Omega \backslash \mathcal{O}.$ 
%%%%%%% robot dynamics %%%%%%%
The \textit{dynamics of the robots} are governed by the following equation
    \begin{align}
        \mbf{p}_i(t+1) = f_i(\mbf{p}_i(t),\mbf{u}_i(t)), \quad i = \{1,\dots,N\} \label{eq:robot_dynamics}
    \end{align}
where $\mbf{p}_i(t) \in \Omega_{\text{free}}$ describes the state of robot $i$ at discrete time $t$ (e.g position), $\mbf{u}_i(t) \in \mcal{U}_i$ is the control input applied to robot $i$ at time $t$ from a \textit{finite} space $\mcal{U}_i$ of admissible control inputs.
%%%%%% hidden states %%%%%%%
The task of the robots is to collaboratively estimate a hidden state $\mbf{x}(t)$ evolving in $\Omega_{\text{free}}$ governed by the following dynamics
    \begin{align}
        \mbf{x}(t+1) = A\mbf{x}(t) + \mbf{w}(t), &\quad \mbf{w}(t) \sim \mathcal{N}(\mbf{d}(t),{\bf{Q}}(t)) \label{eq:target_dynamics}
    \end{align}
where $\mbf{x}(t) \in \mbb{R}^{d_{x}}$ and $\mbf{w}(t) \in \mbb{R}^{d_{{w}}}$ denote the hidden states and the process noise at time $t$, respectively.
%%%%% observation model %%%%%%
Robots are equipped with sensors (e.g., cameras) which allow them to take measurements associated with the unknown state $\mbf{x}(t)$. Hereafter, we assume that the robots can generate measurements as per the following \textit{observation model}
    \begin{align}
        \mbf{y}_i(t) = M_{i}(\mbf{p}_i(t))\mathbf{x}(t) + \mbf{v}_i(t) \label{eq:observation_model}
    \end{align}
where $\mbf{y}_i(t)$ is the measurement signal at time $t$ and $\mbf{v}_i(t) \sim \mathcal{N}(\mbf{0}, \mbf{R}_i(t))$ is a sensor-state dependent measurement noise. Signals observed by robot $i$ are independent of other robots' observations.

\begin{assumption}
The dynamics of the state $\mbf{x}(t)$ in \eqref{eq:target_dynamics}, the control input $\mbf{d}(t)$ the  observation  model \eqref{eq:observation_model},  and  process  and  measurement noise covariances $Q(t)$ and $R_i(t)$ are known for all time instants $t$. This assumption is common in the literature \cite{atanasov2014information}, \cite{atanasov2014joint} and it is required for the application of a Kalman filter to estimate the hidden states.
\end{assumption}
%
%Let $\mbf{p}_{i, 0:t} = [\mbf{p}_i(0)^T, \dots, \mbf{p}_i(t)^T]^T$ and $\mbf{y}_{i,0:t} = [\mbf{y}_i(0)^T,\dots,\mbf{y}^T_i(t)]^T$ denote compactly the evolution of states and measurements signals  of robot $i$ until time $t$. Similarly define $\mbf{u}_{i,0:t} = [\mbf{u}_i(0)^T,\dots,\mbf{u}_i(t)^T]^T$.
%
%%%%%%% conditional pdf %%%%%%%%%
%Due to linearity of the observation model \eqref{eq:observation_model} in the hidden state, the distribution of $\mbf{x}(t)$ given $\mbf{y}_{i,0:t-1}$ remains Gaussian for $t>1$, i.e $\mbb{P}(\mbf{x}(t) \vert \mbf{y}_{i,0:t-1}) = \mcal{N}(\bm{\mu}_i(t\vert \mbf{y}_{i,0:t-1}), \Sigma_i(t\vert \mbf{y}_{i,0:t-1}))$ where $\bm{\mu}_i(t\vert \mbf{y}_{i,0:t-1}$ and $\Sigma_i(t\vert \mbf{y}_{i,0:t-1}))$ denote the a posteriori mean and covariance matrix of $\mbf{x}(t)$ after fusing measurements $\mbf{y}_{i,0:t-1}$ of robot $i$. Note that $\Sigma_i(t)$ can be computed from Bayes Filter which due the to the linear Gaussian assumptions is equivalent to the Kalman filter.

%%%%%%%%%%%%%%%%%%%%%%%%%%%%%%%%%%%%%%%%%%%%%%%%%%%%%%%%%%
%%%%%%%%%%%%%%Communication%%%%%%%%%%%%%%%%%%%%%%%%%%%%%%%
%%%%%%%%%%%%%%%%%%%%%%%%%%%%%%%%%%%%%%%%%%%%%%%%%%%%%%%%%%
\subsection{Distributed Kalman Filter for State Estimation}

%We assume that the robots can communicate with each other and exchange their collected  through an underlying communication network modeled as an undirected graph $G = (V,E)$. The sets $V$ and $E$ denote the set of vertices, defined as $V\coloneqq \{1,\dots, N \}$ and indexed by the robots, and set of edges where existence of an edge $(i,j)$ means that robots $i$ and $j$ can directly communicate with each other. 

%Hereafter, we assume that the hidden state $\bbx(t)$ follows a Gaussian distribution for all times $t\geq 0$, i.e., $\bbx(t)\sim\ccalN(\boldsymbol\mu(t|\bby_{i,0:t}),\Sigma(t|\bby_{i,0:t}))$, where $\bby_{i,0:t}$ denote the measurements that robot $i$ has collected until time $t$, and $\boldsymbol\mu(t|\bby_{i,0:t})$ and $\Sigma(t|\bby_{i,0:t})$ denote the a-posteriori mean and covariance matrix of $\bbx(t)$, respectively, after fusing measurements $\bby_{i,0:t}$. 
Given a Gaussian prior distribution for $\mbf{x}(0)$, i.e., $\mbf{x}(0)\sim\mathcal{N}(\boldsymbol\mu(0),\Sigma(0))$, and measurements, denoted by $\mbf{y}_{i,0:t}$, that robot $i$ has collected until a time instant $t$, robot $i$ computes  a Gaussian distribution that the hidden state follows at time $t$, denoted by $\hat{\mbf{x}}_i(t) = \mathcal{N}(\bm{\mu}_i(t|\mbf{y}_{i,0:t}), \Sigma_i(t|\mbf{y}_{i,0:t}))$, where $\bm{\mu}_i(t|\mbf{y}_{i,0:t})$  and $\Sigma_i(t|\mbf{y}_{i,0:t})$ denote the a-posteriori mean and covariance matrix. To compute this local Gaussian distribution, we adopt the \textit{Distributed Kalman Filter} (DKF) algorithm proposed in \cite{atanasov2014joint}. 
To this end, we assume that the robots formulate an underlying communication network modeled as an undirected graph $G = (V,E)$. The sets $V$ and $E$ denote the set of vertices, defined as $V\coloneqq \{1,\dots, N \}$ and indexed by the robots, and set of edges where existence of an edge $(i,j)$ means that robots $i$ and $j$ can directly communicate with each other. Given, such a communication graph, every robot $i$ updates its respective Gaussian distribution as follows:
    \begin{equation}
    \Omega_i(t+1) = \sum\limits_{j \in \mathcal{N}_i \cup \{i\}} \kappa_{ij} \Omega_j(t) + M_i(t)^T R_i(t)^{-1} M_i(t) \label{eq:distributed_kalman_filter}
    \end{equation}
where $\Omega_i(t) = \Sigma_i(t)^{-1}$, $\mathcal{N}_i$ is the set of nodes (neighbors) connected to robot $i$, $\kappa_{ij} > 0, \sum_{j \in \mathcal{N}_i \cup  \{i\}} \kappa_{ij} = 1$.
%\begin{assumption}
%The graph $G$ modeling the communication network of the robots is assumed be a connected graph $G$ at all times. \textcolor{red}{This assumption is required for... (is this a requirement for the distributed kalman filter?)}\textcolor{blue}{$\leftarrow$ this assumption is requested in DKF to guarantee convergence of $\hat{x}_i \rightarrow x, \forall i \in \{1,\dots,N\}$. So for moving targets I believe is needed!}
%\end{assumption}

%%%%%%%%%%%%%%%%%%%%%%%%%%%%%%%%%%%%%%%%%%%%%%%%%%%%%%%%%%%%%%%%%%%%%%%%%%%%%%%%%%%%%%%%%%%%%%%%%%%%%%%%%%%%%%%%%%
%%%%%Problem Definition %%%%%%%%%%%%%%%%%%%%%%%%%%%%%%%%%%%%%%%%%%%%%%%%%
\subsection{Active Information Acquisition}
The quality of measurements taken by robot $i$ up to a time instant $t$, $\mbf{y}_{i,0:t}$, can be evaluated  using  information  measures,  such  as  the mutual information between $\mbf{y}_{i,0:t}$ and $\mbf{x}(t)$ or the conditional entropy of $\mbf{x}(t)$ given $\mbf{y}_{i,0:t}$. 
%The information available to robot $i$ at time $t$ to compute ${\bf{u}}_i(t)$ is
%\begin{align}
%    \mcal{I}_i(0) = \mbf{y}_i(0), \quad \mathcal{I}_i(t) = (\mathbf{y}_{i,{0:t}}, \mathbf{u}_{i,{0:t-1}}) \label{eq:info_available}
%\end{align}
%%%%%%%%%%%%%%%%%%%%%%%%%%%%%%%%%%%%%%%%%%%%%%%%%%%%%%%%%%%%%%%%%%%%%%%%%%%%
%%%%%%%%%%Stochastic Control%%%%%%%%%%
%%%%%%%%%%%%%%%%%%%%%%%%%%%%%%%%%%%%%%%%%%%%%%%%%%%%%%%%%%%%%%%%%%%%%%%%%%%%
Given initial robot states $\mbf{p}_i(0) \in \Omega_{\text{free}}$ and a prior distribution of the hidden states $\mbf{x}(0)$, our goal is to compute a sequence of control policies $\bm{\pi}_i(t) \colon \{\mbf{u}_{i,0:t-1}, \mbf{y}_{i,0:t} \} \rightarrow \mbf{u}_i(t) \in \mcal{U}_i$ and a planning horizon $F$, for all robots $i \in \{1,\dots, N\}$ and time instants $t=\{0,\dots,F\}$ which
solves the following stochastic optimal control problem:%\footnote{{\textcolor{red}{I think it'd be a bit more understandable if we stated the stochastic problem as a minimization problem with the `determinant' instead of mutual information. I think we will not use mutual info in the rest of the paper. So let's remove this notation/definition :)}}\textcolor{blue}{discussed}}
\begin{subequations}
\label{eq:Prob1}
\begin{align}
& \min\limits_F \max\limits_{\bm{\pi}_{0:F}}\left[J(F,\bm{\pi}_{0:F}) = \sum_{i=1}^{N}\sum_{t=0}^F \mathbb{I}(\mathbf{x}(t+1); \mathbf{y}_{i,{0:t}})\right] \label{eq:obj1}\\
& \ \ \ \ \ \ \mbb{I}(\mbf{x}(F+1);\mbf{y}_{i,0:F}) \geq \epsilon \textit{ for at least one robot},  \label{eq:constr11} \\
& \ \ \ \ \ \    \mbf{p}_i(t+1) = f_i(\mbf{p}_i(t),\bm{\pi}_i(t)), \label{eq:constr12}\\
& \ \ \ \ \ \   \mbf{p}_i(t+1) \in \Omega_{\text{free}}^N, \label{eq:constr13} \\
& \ \ \ \ \ \    \mbf{y}_i(t) = M_i(\mbf{p}_i(t)) \mbf{x}(t) + \mbf{v}_i(t),\label{eq:constr14} \\
& \ \ \ \ \ \    
\mbf{x}(t+1) = A \mbf{x}(t) + \mbf{w}(t) \label{eq:constr15}
\end{align}
\end{subequations} 
where the objective in \eqref{eq:obj1} captures the accumulated mutual information between the measurements received up to time $t$ and the state $\mbf{x}(t+1)$ of all robots and $\bm{\pi}_{0:F}$ stands for the concatenated sequence of control inputs applied to the robots from $t=0$ until $t=F$. The constraint in \eqref{eq:constr11} requires that at the end of the planning there exists at least one robot that has a comprehensive view of the state $\mbf{x}(F+1)$ with a certain confidence $\epsilon \geq 0$. Moreover not including constraint \eqref{eq:constr11} would result in robots to stay put, i.e $F=0$. The constraints \eqref{eq:constr12}, \eqref{eq:constr14} capture the robot and sensor model, respectively, while constraint \eqref{eq:constr15} is the hidden state model. Obstacle avoidance is ensured through constraint \eqref{eq:constr13}.
%\textcolor{red}{I'd delete the rest of the blue text since it's more about the solution and not the problem formulation}\textcolor{blue}{it's something that George mentioned on the chat.. should i keep it?}
%Each robot $i$ aims to address problem \eqref{eq:Prob1} by solving the interior summation of the objective function\footnote{\textcolor{red}{'by solving the interior summation of the objective function': I think this is somewhat ambiguous because the robots try to solve (5) collaboratively. But I still think this paragraph is related to the solution. I would remove it from here. I have added a remark after you define (6).  }} \eqref{eq:obj1}, subject to its own dynamics and measurement model \eqref{eq:constr12} and \eqref{eq:constr13} while ensuring the constraint \eqref{eq:constr11} is satisfied. The problem further requires robot $i$ to inform its neighbors $\mcal{N}_i$ of whether or not the constraint \eqref{eq:constr11} is satisfied on its behalf to ensure that a common horizon $F$ is picked.

Active Information Problem \eqref{eq:Prob1} is a stochastic optimal control problem for which closed-loop policies outclass open loop ones. Nonetheless, the linear relation between measurement $\mbf{y}_i(t)$ and state $\mbf{x}(t)$ in \eqref{eq:observation_model} and the gaussian assumptions transform the stochastic optimal control problem \eqref{eq:Prob1} into a deterministic optimal control problem where open loop policies are optimal \cite{atanasov2014information}. The principle is extended for our case where robots communicate with neighbors $\mcal{N}_i$ according to a communication graph $G = (V,E)$ and fuse their beliefs via DKF \eqref{eq:distributed_kalman_filter} and an open-loop control sequence $\sigma = \mbf{u}(0),\dots,\mbf{u}(F)$ exists which is optimal in \eqref{eq:Prob1}.

%\begin{theorem}[Separation Principle]
%\label{thm:separation_principle}
%\textit{If the prior distribution of $\mbf{x}(0)$ given measurements $\mbf{y}_{i}(0)$ is gaussian with covariance matrix $\Sigma_i(0 \vert \mbf{y}_i(0))$ and robots communicate with neighbors $\mathcal{N}_i$ according to a communication graph $G=(V,E)$, there exists an open-loop control sequence $\sigma = \mbf{u}(0),\dots,\mbf{u}(F)$ which is optimal in \eqref{eq:Prob1}}. 
%\end{theorem} 
Furthermore Problem \eqref{eq:Prob1} can be transformed to the following deterministic control problem 
%%%%%%%%%%%%%%%%%%%%%%%%%%%%%%%%%%%%%%%%%%%%%%%%%%%%%%%%%%%%%%%%%%%%%%%%%%%%
%%%%%Final Optimization problem%%%%%%%
%%%%%%%%%%%%%%%%%%%%%%%%%%%%%%%%%%%%%%%%%%%%%%%%%%%%%%%%%%%%%%%%%%%%%%%%%%%%
\begin{subequations}
\label{eq:Prob2}
\begin{align}
& \min_{\substack{ F, \bm{u}_{0:F}}} \left[J(F,\bm{u}_{0:F}) = \sum_{i=1}^{N}\sum_{t=0}^{F}  \det\Sigma_i(t+1\vert \mathbf{y}_{i,{0:t}}) \right] \label{eq:obj2}\\
& \ \ \ \ \ \ \ \det\Sigma_i(F+1)\leq \delta(\epsilon) \textit{ for at least one robot}  \label{eq:constr21} \\
& \ \ \ \ \ \ \ \    \mbf{p}_i(t+1) \in \Omega_{\text{free}}^N, \label{eq:constr22} \\
& \ \ \ \ \ \ \ \  \mbf{p}_i(t+1) = f_i(\mbf{p}_i(t),\mbf{u}_i(t)), \label{eq:constr23}\\
& \ \ \ \ \ \ \ \ \Omega_i(t+1) =\rho(\mbf{p}_i(t),\Omega_{j \in \mathcal{N}_i \cup \{i\}}(t)),\label{eq:constr24} \\
& \ \ \ \ \ \ \ \  \Sigma_i(t+1) = \Omega_i(t+1)^{-1} \label{eq:constr25}
\end{align}
\end{subequations} 
where $\rho(\cdot)$ is the DKF update rule in \eqref{eq:distributed_kalman_filter}, $\bm{u}_{0:F}$ is the concatenated sequence of control inputs of all robots from $t=0$ until $t=F$ and $\delta(\epsilon) \geq 0$ is the resulting uncertainty threshold from the definition of mutual information and \eqref{eq:constr11}. The robots collaboratively solve the objective \eqref{eq:obj2} with each constraint applying individually to them. Note that solving \eqref{eq:Prob2} requires the robots to exchange their covariance matrices over the communication network $G$ due to \eqref{eq:constr24}.
%Each robot $i$ aims to address problem \eqref{eq:Prob1} by solving the interior summation of the objective function\footnote{\textcolor{red}{'by solving the interior summation of the objective function': I think this is somewhat ambiguous because the robots try to solve (5) collaboratively. But I still think this paragraph is related to the solution. I would remove it from here. I have added a remark after you define (6).  }} \eqref{eq:obj1}, subject to its own dynamics and measurement model \eqref{eq:constr12} and \eqref{eq:constr13} while ensuring the constraint \eqref{eq:constr11} is satisfied. The problem further requires robot $i$ to inform its neighbors $\mcal{N}_i$ of whether or not the constraint \eqref{eq:constr11} is satisfied on its behalf to ensure that a common horizon $F$ is picked.
%ote that application of\eqref{eq:distributed_kalman_filter} requires the robots to exchange $\Omega_i(t+1)$ with their neighbors over the communication graph $\mathcal{G}$.
%For a complete proof of Theorem \ref{thm:separation_principle} see Section  \ref{sec:appendix}. 

\section{Distributed Sampling Based Active Information Acquisition}
\label{sec:incremental}
%A centralized solution to \eqref{eq:Prob2} is proposed in \cite{kantaros2019asymptotically}. Nevertheless, a centralized solution can incur a high computational cost as it requires exploring a high dimensional joint space composed of the space of multi-robot states and covariance matrices. As a result, its computational cost increases as the number of robots and/or hidden states increases. 

%To mitigate this issue, we propose 
A novel distributed sampling-based method to solve \eqref{eq:Prob2} %, that builds upon our previous work \cite{kantaros2019asymptotically}
is proposed which is summarized in Algorithm \ref{alg:RRT}. In the proposed method, each robot $i$ builds, in parallel with all other robots, a local directed tree that explores both the information space and its robot motion space while exchanging information with neighboring robots collected in $\mathcal{N}_i$, opposed to the global tree built in \cite{kantaros2019asymptotically} resulting in decreasing the computational complexity significantly.\footnote{Throughout the paper robot $i$ shall be used as a reference. Same procedure is followed for all the robots unless stated differently.} 

%\footnote{\textcolor{red}{I think we need to somehow show this theoretically. For example, we could show the worst-case computational complexity of adding a node in a local tree is less than the corresponding complexity for the global tree.}} 
%
The proposed distributed sampling-based algorithm is presented in Algorithm \ref{alg:RRT}. In what follows, we denote the tree built by robot $i$ as $\mathcal{G}_i = \{\mathcal{V}_i, \mathcal{E}_i, {J_{\mathcal{G}}}_i\}$, where $\mathcal{V}_i$ is the set of nodes and $\mathcal{E}_i$ denotes the set of edges. The set of nodes collects states of the form $\mbf{q}_i(t) = [\mbf{p}_i(t), \Sigma_i(t), \mathcal{S}_i(t)]$ where $\mathcal{S}_i(t)$ is a set that collects nodes $\mbf{q}_j \in \mcal{V}_j$ for $j \in \mcal{N}_i$, that participate in the update rule of $\Sigma_i(t)$ as per \eqref{eq:distributed_kalman_filter} and $[\mcal{S}_i(t)]_j = \mbf{q}_j(t)$, where $[\mcal{S}_i(t)]_j$ denotes the $j$-th element in the set $\mcal{S}_i(t)$; see Section \ref{sec:extending}. %\footnote{\textcolor{red}{We need to briefly and informally explain what it is and say that it will be defined formally later in the text.}\textcolor{blue}{fixed}} 
The root of the tree is defined as $\mbf{q}_i(0) = [\mbf{p}_i(0), \Sigma_i(0), \mcal{S}_i(0)]$ where $\mbf{p}_i(0), \Sigma_i(0)$ are the initial state of the robot and prior covariance, respectively, and $\mcal{S}_i(0) = \emptyset$.
%
%Robot $i$ interacts with its neighbors $\mathcal{N}_i$ and incrementally builds, in parallel to other robots, its own directed tree $\mathcal{G}_i = \{\mathcal{V}_i, \mathcal{E}_i, {J_{\mathcal{G}}}_i\}$, where $\mathcal{V}_i$ is the set of nodes and $\mathcal{E}_i$ denotes the set of edges. States of the form $\mbf{q}_i(t) = [\mbf{p}_i(t), \Sigma_i(t), \mathcal{S}_i(t)]$ are assigned to each node of $\mathcal{V}_i$ of depth $t$.%%footnote%%
 
%%%%%%%%%%%%%%%%%%%%%%%%%%%%%%%%%%%%%%%%%%%
%%%%%%%%%%%% cost function %%%%%%%%%%%%%%%%
%%%%%%%%%%%%%%%%%%%%%%%%%%%%%%%%%%%%%%%%%%%

The cost function ${J_{\mcal{G}}}_i \colon \mcal{V}_i \rightarrow \mbb{R}_{+}$ describes the cost of reaching a node $\mbf{q}_i(t)$ starting from the root of the tree $\mbf{q}_i(0)$. The cost of the root $\mbf{q}_i(0)$ is ${J_{\mcal{G}}}_i(\mbf{q}(0)) = \det \Sigma_i(0)$, while the cost of $\mbf{q}_i(t+1) = [\mbf{p}_i(t+1), \Sigma_i(t+1), \mcal{S}_i(t+1)]$ is equal to
\begin{align}
    {J_{\mcal{G}}}_i(\mbf{q}_i(t+1)) = {J_{\mcal{G}}}_i(\mbf{q}_i(t)) + \det \Sigma_i(t+1)
    \label{eq:cost}
\end{align}
where $\mbf{q}_i(t)$ is the parent node of $\mbf{q}_i(t+1)$. Applying \eqref{eq:cost} recursively results in ${J_{\mcal{G}}}_i(\mbf{q}_i(t+1)) = \sum_{k=0}^{t+1} \det \Sigma_i(k)$ which is equal to the interior sum of the objective in \eqref{eq:Prob2}. Thus, the summation of costs of individual tree paths over all robots yields the objective function in  \eqref{eq:obj2}. %If node $\mbf{q}_i(t+1)$ is summed over all the robots, $i=1,\dots,N$, 

The tree $\mcal{G}_i$ is initialized so that $\mcal{V}_i = \{\mbf{q}_i(0) \}, \ \mcal{E}_i = \emptyset$ and ${J_{\mcal{G}}}_i(\mbf{q}_i(0))=\det \Sigma_i(0)$ [line \ref{rrt:init}, Alg. \ref{alg:RRT}]. The construction of the trees lasts for $n_{\max}$ iterations and is based on two procedures \textit{sampling} [line \ref{rrt:samplekrand}-\ref{rrt:sampleu}, Alg. \ref{alg:RRT}] and \textit{extending-the-tree} [line \ref{rrt:for2}-\ref{rrt:updKn}, Alg. \ref{alg:RRT}]. A solution consisting of a terminal horizon $F$ and a sequence of paths control inputs $\mbf{u}_{0:F}$ is returned after $n_{\max}$ iterations.

\setlength{\textfloatsep}{1pt}
%%%%%%%%%%%%%%%%%%%%%%%%%%%%%%%%%%%%%
%%%%%%%%% Algorithm %%%%%%%%%%%%%%%%%
%%%%%%%%%%%%%%%%%%%%%%%%%%%%%%%%%%%%%
\begin{algorithm}[t]
\footnotesize
\caption{Sampling-based Active Information Acquisition}
\LinesNumbered
\SetKwBlock{Parfor}{parfor $i=1,\dots,N$}{end}
\SetKwBlock{ParFor}{parfor $i \in \mcal{N}^*$}{end}
\label{alg:RRT}
\KwIn{ (i) maximum number of iterations $n_{\text{max}}$, (ii) dynamics \eqref{eq:robot_dynamics}, \eqref{eq:target_dynamics}, observation model \eqref{eq:observation_model}, (iii) prior Gaussians $\mathcal{N}(\bm{\mu}_{i}(0),\Sigma_{i}(0))$, (iv) initial robot configurations $\bm{p}_i(0)$}
\KwOut{Terminal horizon $F$, and control inputs $\bm{u}_{0:F}$}
 Initialize $\mathcal{V}_i = \{\mbf{q}_i(0)\}$, $\mathcal{E}_i = \emptyset$, $\mathcal{V}_{i,1}=\{\bm{q}_i(0)\}$, $K_{i,1}=1$ and $\mathcal{X}_g = \emptyset$;\label{rrt:init} \\
\For{ $n = 1, \dots, n_{\text{max}}$}{\label{rrt:forn}
\Parfor{\label{rrt:fori}
	 Sample a subset $\mathcal{V}_{i,k_{\text{rand}}}$ from $f_{\mathcal{V}_i}$\;\label{rrt:samplekrand}
	 Sample a control input $\bm{u}_{i,\text{new}}\in\mathcal{U}_i$ from $f_{\mathcal{U}_i}$ and compute $\bm{p}_{i,\text{new}}$\;\label{rrt:sampleu}
	\If{$\bm{p}_{i,\text{new}}\in\Omega_{\text{free}}^N$}{\label{rrt:obsFree}
		\For{$\bm{q}_{i,\text{rand}}(t)=[\bm{p}_{i,\text{rand}}(t),\Sigma_{i,\text{rand}}(t), \mathcal{S}_{i, \text{rand}}(t)]\in\mathcal{V}_{i,k_{\text{rand}}} $}{\label{rrt:for2}
		Compute sets $\mcal{Q}_{ij}(t+1)$ for all $j\in\mcal{N}_i$ \; \label{rrt:Q}
		Sample node $\bm{q}_j$ from $f_{\mcal{Q}_i}$ for all $j\in\mcal{N}_i$ and compute $\mcal{S}_{i,\text{new}}$\; \label{rrt:Snew}
		 Compute $\Sigma_{i,\text{new}}(t+1)=\rho(p_{i,\text{rand}}(t),\Sigma_{i,\text{rand}}(t), \Sigma_j(t))$\;\label{rrt:sigma_new}
		 Construct $\bm{q}_{i,\text{new}}(t+1)=[\bm{p}_{i,\text{new}}(t+1),\Sigma_{i,\text{new}}(t+1), \mcal{S}_{i,\text{new}}]$\;\label{rrt:q_new}

		 Update set of nodes: $\mathcal{V}_i= \mathcal{V}_i\cup\{\bm{q}_{i,\text{new}}\}$\;\label{rrt:updV}
		 Update set of edges: $\mathcal{E}_i = \mathcal{E}_i\cup\{(\bm{q}_{i,\text{rand}},\bm{q}_{i,\text{new}})\}$\;\label{rrt:updE}
		 Compute cost of new state: $J_{\mathcal{G}_i}(\bm{q}_{i,\text{new}})=J_{\mathcal{G}_i}(\bm{q}_{i,\text{rand}})+\det \Sigma_{i,\text{new}}(t+1)$ \label{rrt:updCost}

		\If{$\exists k\in\{1,\dots,K_n\}$ associated with same position as in $\bm{q}_{i,\text{new}}$}{\label{rrt:if2}
		 $\mathcal{V}_{i,k}=\mathcal{V}_{i,k}\cup\{\bm{q}_{i,\text{new}}\}$\;}\label{rrt:updVk}
		\Else{
		 $K_{i,n}=K_{i,n}+1$, $\mathcal{V}_{i,K_n}=\{\bm{q}_{i,\text{new}}\}$\;\label{rrt:updKn}
		}
		\If{$\bm{q}_{\text{new}}$ satisfies
		\eqref{eq:constr21}}{\label{rrt:updGoal1}
			 ${\mathcal{X}_g}_i = {\mathcal{X}_g}_i \cup\{\bm{q}_{\text{new}} \} $\;\label{rrt:updGoal2} 
		}}}}}
\ParFor{\label{rrt:parfor2}
\For{$\mbf{q}_i(t_k) \in {\mcal{X}_g}_i$}{\label{rrt:forXgi}
Compute and transmit path $\mbf{q}_{i,0:t_k} = [\mbf{q}_i(0), \dots, \mbf{q}_i(t_k)]$\; \label{rrt:qipaths}
Receive corresponding paths $\mbf{q}_{j,0:t_k}$\;\label{rrt:qjpaths}
Compute cost $J(t_k)$ as per \ref{eq:obj2}\;
}\label{rrt:totalcost}
Compute minimum cost $J({t_{\text{end}}})$\; \label{rrt:mincost}
}
Set $F=t_{\text{end}}$ and execute team paths $\mbf{q}_{0:t_{\text{end}}}$\;\label{rrt:final}

%\Parfor{ \label{rrt:parfor2}
%Compute and transmit smallest cost $J_{\mcal{G}_i}(t_{i, \text{end}})$\; \label{rrt:costlocal}
%Initialize $\mcal{J}_i = \{J_{\mcal{G}_i}(t_{i, \text{end}})\}$\; \label{rrt:initJ} 
%\While{$t_{\text{wait}}$}{\label{rrt:while}
%Receive and transmit $J_{\mcal{G}_j}(t_{j, \text{end}})$\;
%Update set $\mcal{J}_i = \mcal{J}_i \cup J_{\mcal{G}_j}(t_{j,\text{end}})$\;
%\label{rrt:updateJ}
%}
%Determine optimal robot $k = \arg \min \mcal{J}_i$\; \label{rrt:finalrobot}
%}
%$F = t_{k, \text{end}}$ and compute paths $\mbf{q}^*_{0:F}$\;\label{rrt:finalsol}
\end{algorithm}

To extract such a solution, we need first to define the goal set $\mcal{X}_{g_i}\subseteq\mcal{V}$ that collects all states $\mbf{q}_i(t) = [\mbf{p}_i(t), \Sigma_i(t), \mcal{S}_i(t)]$ in the $i$-th tree that satisfy $\det \Sigma_i(t) \leq \delta$ [lines \ref{rrt:updGoal1}-\ref{rrt:updGoal2}, Algorithm \ref{alg:RRT}]. After $n_{\max}$ iterations, every robot $i$ with non-empty goal set, collected in the set $\mcal{N}^*$, selects a node in $\mcal{X}_{g_i}$ and computes %node $\mbf{q}_{t_{i,\text{end}}} \in {\mcal{X}_g}_i$ with the smallest cost, i.e $\mbf{q}_{t_{i,\text{end}}} = \arg \\ \min_{\mbf{q} \in {\mcal{X}_g}_i} \det \Sigma_i$ and transmits the cost 
the path connecting it to the root of its tree. This path is then propagated to all robots in the network in a multi-hop fashion through the communication graph $G$. % $\mcal{N}_i$. 
Using the sets $\mcal{S}_i(t)$, all robots compute a path in their trees that they should follow so that robot $i\in\mcal{N}^*$ can follow its selected/transmitted path. The resulting paths are then transmitted back to robot $i$ which computes the total cost as per \eqref{eq:obj2}. This process is repeated for all nodes in $\mcal{X}_{g_i}$ and for all robots $i$ in $\mcal{N}^*$, [lines \ref{rrt:forXgi}-\ref{rrt:totalcost}, Alg. \ref{alg:RRT}]. Among all resulting team paths, the robots select the team path with the minimum cost as \eqref{eq:obj2}, [lines \ref{rrt:mincost}-\ref{rrt:final}, Alg. \ref{alg:RRT}]. 

%%%%%%%%%%%%%%%%%%%%%%%%%%%%%%%%%%%%%%%%%%%%%%
%%%%%% Incremental Construction of trees %%%%%
%%%%%%%%%%%%%%%%%%%%%%%%%%%%%%%%%%%%%%%%%%%%%%
\subsection{Incremental Construction of trees}
At each iteration $n$ a node $\qrand \in \mcal{V}_i$ is selected to be expanded according to a sampling procedure described in Section \ref{sec:sampling}. A new node $\qnew = \left[\pnew, \Sigmanew, \mcal{S}_{i,\text{new}}\right]$ is added to $\mcal{V}_i$ and the edge $(\qrand, \qnew)$ is added to $\mcal{E}_i$; see Section \ref{sec:extending}.

\begin{figure}[t]
     \centering
     \begin{subfigure}[b]{0.23\textwidth}
         \centering
         \includegraphics[width=0.42\textwidth]{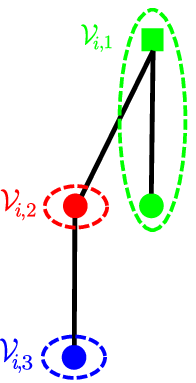}
         \caption{}
         \label{fig:sampling_example}
     \end{subfigure}
     \hfill
     \begin{subfigure}[b]{0.23\textwidth}
         \centering
         \includegraphics[width=0.65\textwidth]{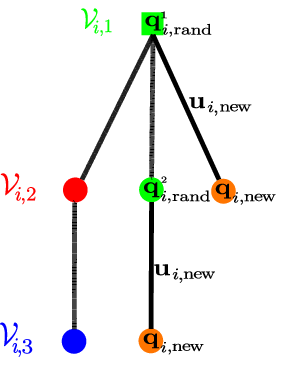}
         \caption{}
         \label{fig:extending_example}
     \end{subfigure}
     \caption{Fig. \ref{fig:sampling_example} shows an example of the sets $\mcal{V}_{i,k}$. The root of the tree is depicted by a square. There are three ($K_{i,n} = 3)$ groups $\mcal{V}_{i,k}$ and the  members of each group have the same color. Group $\mcal{V}_{i,1}$ consists of nodes $\mbf{q}_i^1$ and $\mbf{q}_i^2$ for which $\mbf{p}_i^1 = \mbf{p}_i^2$. Figure \ref{fig:extending_example} illustrates the incremental construction of the tree.}
     \label{fig:construction_of_tree_example}
\end{figure}

%%%%%%%%%%%%%%%%%%%%%%%%%%%%%%%%
%%%%%% Sampling Strategy %%%%%%%
%%%%%%%%%%%%%%%%%%%%%%%%%%%%%%%%
\subsubsection{\textbf{Sampling Strategy}}\label{sec:sampling}
The sampling procedure begins by first dividing the set of nodes $\mcal{V}_i$ into a \textit{finite} number of sets $\mcal{V}_{i,k} \subseteq \mcal{V}_i$. If $K_{i,n}$ is the number of created sets during iteration $n$ then $\mcal{V}_i = \mathop{\cup}_{k=1}^{K_{i,n}} \mcal{V}_{i,k}$. The set $\mcal{V}_{i,k}$ collects nodes $\mbf{q}_i = [\mbf{p}_i, \Sigma_i, \mcal{S}_i] \in \mcal{V}_i$ that share the same robot state $\mbf{p_i}$; see Fig. \ref{fig:sampling_example}. At iteration $n=1$ it holds that $K_{i,1} = 1$ and $\mcal{V}_i = \mcal{V}_{i,1}$ [line \ref{rrt:init}, Alg. \ref{alg:RRT}]. Once the sets $\mcal{V}_{i,k}$ are created, the nodes $\mbf{q}_{i,\text{rand}} \in \mcal{V}_{i,k_{i,\text{rand}}}$ will be expanded where index $k_{i,\text{rand}}$ is sampled from a given distribution $f_{\mcal{V}_i}(k_i\vert \mcal{V}_i) \colon \{1,\dots, K_{i,n}\} \rightarrow [0,1]$ and points to the set $\mcal{V}_{i, k_{i,\text{rand}}}$[line \ref{rrt:samplekrand}, Alg. \ref{alg:RRT}].

Given nodes $\mbf{q}_{i,\text{rand}}$ to be expanded  with corresponding robot state $\prand$, a control input $\unew$ is sampled from a distribution $f_{\mcal{U}_i}(\mbf{u}_i) \colon \mcal{U}_i \rightarrow [0,1]$. The new robot state $\pnew$ is created by applying the selected control input $\unew$ to $\prand$ according to \eqref{eq:robot_dynamics} [line \ref{rrt:sampleu}, Alg. \ref{alg:RRT}]. 

\begin{remark}[Diversity of trees]
\label{rem:diversity}
The selection of which group $\mcal{V}_{i,k}$ is expanded is  independent between different robots. Due to the randomness imposed by the aforementioned sampling procedures, the constructed trees $\mcal{G}_i$ at iteration $n$ may have different structures among the robots.
\end{remark}
%%%%%%%%%%%%%%%%%%%%%%%%%%%%%%%
%%%%%%% Extending the tree %%%%
%%%%%%%%%%%%%%%%%%%%%%%%%%%%%%%

\subsubsection{\textbf{Extending the tree}}\label{sec:extending}
\begin{figure}[b]
    \centering
    \includegraphics[width=0.38\textwidth]{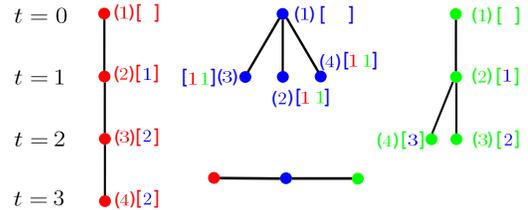}
    \setlength{\belowcaptionskip}{-10pt}
    \caption{This Figure illustrates the computation of $\Sigmanew$ and $\mcal{S}_{i,\text{new}}$ for a group of three robots, $i \in $ \{red, blue, green\}. The communication graph is depicted at the bottom. The growth of the trees for four iterations is presented where node $(k)$ is added to the trees. In square brackets, sets $\mcal{S}_{i}$ are described. Red node $(4)$ awaits information from the blue robot, while the parent red node $(3)$ transmits its own $\mcal{S}_{\text{red}}$, i.e. informs that it once communicated with the blue node $(2)$. Blue node $(2)$ is a leaf therefore $\mcal{Q}_{\text{red}} = \{2\}$ and so $\mcal{S}_{\text{red}, \text{new}} = [2]$. Similarly for green node $(4)$, $\mcal{Q}_{\text{green}} = \{2,3,4 \}$ and blue node $(3)$ is sampled from $f_{\mathcal{Q}}$ to constitute $\mcal{S}_{\text{green}, \text{new}}$.
    }
    \label{fig:sigma_new}
\end{figure}

Once $\pnew$ is constructed, we check whether it belongs to the obstacle-free area $\Omega_{\text{free}}$ or not. In the latter case, the sample is rejected and the sampling procedure is repeated [line \ref{rrt:obsFree}, Alg. \ref{alg:RRT}]. Otherwise, 
the state $\qnew$ is constructed; see Fig. \ref{fig:extending_example}. Specifically, given the parent node $\qrand = [\prand, \Sigmarand, \mcal{S}_{i, \text{rand}}]$ and $\pnew$, the construction of $\qnew$ requires the computation of $\Sigmanew$ and $\mcal{S}_{i, \text{new}}$. %Assume that $t$ is the depth of $\qrand$ and $t+1$ the depth of $\qnew$.
To what follows we assume that $\qrand$ and $\qnew$ lie at depth $t$ and $t+1$ respectively.\footnote{To simplify notation we drop the \textit{rand} and \textit{new} subscripts since the relation breaks down to parent and child node where depths $t$ and $t+1$ discriminates them.}
The update rule in \eqref{eq:distributed_kalman_filter} requires covariance matrix $\Sigma_i(t)$ obtained from node $\mbf{q}_i(t)$ and $\Sigma_j(t)$ from neighbors $j \in \mcal{N}_i$. From Remark \ref{rem:diversity}, at depth $t$ there may exist more than one nodes $\mbf{q}_j(t) \in \mcal{V}_j$  to provide $\Sigma_j(t)$ or no nodes. Moreover, the covariance matrix $\Sigma_i(t)$ of node $\mbf{q}_i (t)$ was once constructed using information received by the set of nodes $\mcal{S}_{i}(t) = \{\mbf{q}_j(t-1) \mid \Sigma_j(t-1) \textit{\text{ participated in computation of }} \Sigma_i(t) \textit{\text{ in \eqref{eq:distributed_kalman_filter}}}, \forall \ j \in \mcal{N}_i \}$, with $\vert \mcal{S}_i(t) \vert = \vert \mcal{N}_i \vert$ and $[\mcal{S}_i(t)]_j = \mbf{q}_j(t-1)$. To maintain consistency with the update rule in \eqref{eq:distributed_kalman_filter} only children of the set of nodes $\mcal{S}_i(t)$ are allowed to distribute their corresponding covariance matrices. For every robot $j \in \mcal{N}_i$ the set of admissible nodes to share $\Sigma_j(t)$ with robot $i$ is $\mathcal{Q}_{ij}(t) = \{\mbf{q}_j(t) \mid \mbf{q}_j(t) \textit{ is child of node } [\mcal{S}_i(t)]_j \}$. Given the set $\mathcal{Q}_{ij}(t)$ node $\tilde{\mbf{q}}_j(t)$ is sampled from a given discrete distribution $f_{\mathcal{Q}_i}(\mbf{q}_j) \colon \mathcal{Q}_{ij}(t) \rightarrow [0,1]$ and the set $\mcal{S}_i(t+1)$ is created by collecting the sampled nodes from all $j \in \mcal{N}_i$ [lines \ref{rrt:Q}-\ref{rrt:Snew}, Alg. \ref{alg:RRT}]; see also Fig. \ref{fig:sigma_new}. The covariance matrix $\Sigma_i(t+1)$ is then computed by applying the update rule in  \eqref{eq:distributed_kalman_filter} and node $\qnew(t+1)$ is completed [lines \ref{rrt:sigma_new}-\ref{rrt:q_new}, Alg. \ref{alg:RRT}]. 

Next, the set of nodes and edges are updated and the cost of node $\qnew$ is computed as $J_{\mcal{G}_i}(\qnew(t+1)) = J_{\mcal{G}_i}(\qrand(t)) + \det \Sigmanew(t+1)$ [lines \ref{rrt:updV}-\ref{rrt:updCost}, Alg. \ref{alg:RRT}]. Finally, the sets $\mcal{V}_{i,k}$ are updated, so that if there already exists a subset $\mcal{V}_{i,k}$ with the same configuration as the state $\pnew$, then $\mcal{V}_{i,k} = \mcal{V}_{i,k}\cup \{\qnew(t+1) \}$. Otherwise, a new subset $\mcal{V}_{i,k} = \{\qnew \}$ is created and the number of subsets increases by one, i.e. $K_{i,n} = K_{i,n} + 1$ [lines \ref{rrt:updVk}-\ref{rrt:updKn}, Alg. \ref{alg:RRT}].

%In what follows, $\mathcal{L}_i^n(t)$ is the set of nodes $\mbf{q}_i \in \mathcal{V}_i$ that lie in depth less than or equal to $t$ and $\mathcal{SC}_i^n(\mbf{q})$ is the set of nodes $\mbf{q}_i \in \mcal{V}_i$ that are subchildren of node $\mbf{q}$ at iteration $n$. For every robot $j \in \mcal{N}_i$ the set of admissible nodes to transmit $\Sigma_j(t)$ to robot $i$ is $\mathcal{Q}_{ij}(t) = \{\mbf{q}_j \mid \mbf{q}_j \in \mcal{V}_j \ \wedge \ \mbf{q}_j \in \mathcal{L}_j^n(t-1) \cap \mathcal{SC}_j^n(\tilde{\mbf{q}}_j(t-1)) \}$. Note that $\mbf{q}_j$ may not lie at depth $t$. Given the sets $\mathcal{Q}_{ij}$ we sample node $\tilde{\mbf{q}}_j(t)$ from a given discrete distribution $f_{\mathcal{Q}_i}(\tilde{\mbf{q}}_j) \colon \mathcal{Q}_{ij} \rightarrow [0,1]$ and the set $\mcal{S}_i(t+1)$ collects the sampled nodes for all $j \in \mcal{N}_i$; see Figure \ref{fig:sigma_new} for an illustrative example.

\begin{remark}[Communication]
The update of $\Sigmanew$ in Section \ref{sec:extending} is the only procedure for which robots need to communicate with each other to build their trees. The information robot $i$ sends to neighbors $\mcal{N}_i$ is the set $\mcal{S}_{i, \text{rand}}$. In exchange it receives covariance matrices $\Sigma_j$.
\end{remark}

\begin{remark}[Set of candidate nodes]
In the case where the set of candidate nodes $\mcal{Q}_{ij}(t)$ is empty, covariance matrix of node $[\mcal{S}_i(t)]_j$ will be transmitted. 
\end{remark}

\section{Complexity, Completeness and Optimality}
\label{sec:completeness}
In this section, we show that Algorithm \ref{alg:RRT} reduces the computational complexity per iteration of its centralized counterpart \cite{kantaros2019asymptotically}. We further show that it is probabilistically complete and asymptotically optimal under the following assumptions regarding the mass functions $f_{\mcal{V}_i}, f_{\mcal{U}_i}, f_{\mcal{Q}_i}$. The proofs of the following results can be found in Section \ref{sec:appendix}.%The proofs of Theorems \ref{thm:probCompl}, \ref{thm:asOpt} \& \ref{thm:time} can be found in Section \ref{sec:appendix}.

\begin{assumption}[\textbf{Probability mass function} $f_{\mcal{V}_i}$] \label{as:fv} (i) Pr-\\obability mass function $f_{\mcal{V}_i}(k_i\vert \mcal{V}_i)  \colon \{1,\dots, K_{i,n} \} \rightarrow [0,1]$ satisfies $f_{\mcal{V}_i}(k_i\vert \mcal{V}_i) \geq \epsilon, \forall k_i \in \{1,\dots,K_{i,n} \}, \forall n \geq 0$ for some $\epsilon > 0$ that remains constant across all iterations (ii) $k_{i,\text{rand}}$ are drawn independently across iterations.
\end{assumption}

\begin{assumption}[\textbf{Probability mass function} $f_{\mcal{U}_i}$]
(i) Pr-\\obability mass function $f_{\mcal{U}_i}$ satisfies $f_{\mcal{U}_i}(\mbf{u}_i) \geq \zeta, \forall \mbf{u}_i \in \mcal{U}_i, \forall n \geq 0$, for some $\zeta >0$ that remains constant across all iterations (ii) Samples $\mbf{u}_i$ are drawn independently across iterations.
\label{as:fu}
\end{assumption}

\begin{assumption}[\textbf{Probability mass function} $f_{\mcal{Q}_i}$]
\label{as:fq} (i) Pr-\\obability mass function $f_{\mcal{Q}_i}$ satisfies $f_{\mcal{Q}_i}(\mbf{q}_j) \geq \xi, \forall \mbf{q}_j \in \mcal{Q}_{ij}, \forall n \geq 0$, where $\xi >0$ (ii) Samples $\mbf{q}_j$ are drawn independently across iterations.
\end{assumption}

\begin{remark}[\textbf{Mass functions}]
Assumptions \ref{as:fv}-\ref{as:fq} are defined in such a way to ensure that the sets (i) $\{1,\dots,K_{i,n} \}$, (ii) $\mcal{U}_i$ and (iii) $\mcal{Q}_{ij}$ have non-zero measure. They are important to ensure the exploration of the entire physical and reachable information space and the fusion of potential beliefs for the DKF. All of them are very flexible, since they allow $f_{\mcal{V}_i}, f_{\mcal{U}_i}$ and $f_{\mcal{Q}_i}$ to change with iterations of Algorithm \ref{alg:RRT}, as the trees grow and to be different among the robots. Any mass functions could be used as long as they satisfy the aforementioned assumptions. A typical example is the discrete uniform distribution. In Section \ref{sec:biased} non-uniform distributions are suggested for a target-tracking scenario.
\end{remark}

\begin{theorem}[\textbf{Probabilistic Completeness}]\label{thm:probCompl}
\textit{If there exists a solution to Problem \ref{eq:Prob2}, then Algorithm \ref{alg:RRT} is probabilistically complete, i.e., feasible paths $\bm{q}_{i,0:F} = \bm{q}_i(0), \dots, \bm{q}_i(F)$, $\bm{q}_i(f)\in\mcal{V}_i$, for all $f\in\{0,\dots,F\}$ and $i\in\{1,\dots,N \}$, will be found with probability $1$, as $n\to\infty$.} 
\end{theorem}

\begin{theorem}[\textbf{Asymptotic Optimality}]\label{thm:asOpt}
\textit{Assume that there exists an optimal solution to Problem \ref{eq:Prob2}. Then, Algorithm \ref{alg:RRT} is asymptotically optimal, i.e., the optimal paths 
$\bm{q}_{i,0:F}^* =  \bm{q}_i(0), \bm{q}_i(1), \bm{q}_i(2), \dots, \bm{q}_i(F)$ for all $i=\{1,\dots,N\}$, will be found with probability $1$, as $n\to\infty$.} 
\end{theorem}

\begin{theorem}[\textbf{Complexity Per Iteration}]\label{thm:time}
\textit{Let $d_{\max}$ denote the maximum degree of the vertices of the communication graph $G$, $N$ be the number of robots and $k$ be the number of nodes $\mbf{q}_{\text{rand}}$ to be expanded in one iteration. Then the computational complexity per iteration of Algorithm $\ref{alg:RRT}$ and the centralized approach \cite{kantaros2019asymptotically} is $\mcal{O}(k d_{\max})$ and $\mcal{O}(k N)$ respectively, under a worst-case scenario complexity analysis.} 

%\textit{Let $d_{\max}$ \\denote the maximum degree of the vertices of the communication graph $G$ and the computational gain of Algorithm \ref{alg:RRT} over the centralized approach in \cite{kantaros2019asymptotically} describe the ratio between the time needed to complete one iteration of Algorithm \ref{alg:RRT} to the time of \cite{kantaros2019asymptotically}. Then the computational gain with parameters $d_{\max}$ and number of robots $N$ is $d_{\max}/N$.}

%\textit{Let $d_{\max}$ \\denote the maximum degree of the vertices of the communication graph $G$. Then the computational gain  per iteration of Algorithm \ref{alg:RRT} over the centralized approach \cite{kantaros2019asymptotically} with parameters $d_{\max}$ and number of robots $N$ is $d_{\max}/N$.} %under a worst-case scenario complexity analysis.}
\end{theorem}

\section{Biased Sampling - Target Tracking}
\label{sec:biased}
In large-scale estimation problems where a large number of robots are equipped to estimate high-dimensional hidden states, mass functions $f_{\mcal{V}_i},f_{\mcal{U}_i}$ and $f_{\mcal{S}_i}$ introduced in Section \ref{sec:incremental} could be designed such that the trees tend to explore regions that are expected to be informative. In this section, we consider an application to target localization for Algorithm \ref{alg:RRT}. The hidden state $\mbf{x}(t)$ now collects the positions of all targets at time $t$, i.e $\mbf{x}(t) = [\mbf{x}_1(t)^T,\dots,\mbf{x}_M(t)^T]^T$, where $\mbf{x}_l(t)$ is the position of target $l$ at time $t$ and $M$ is the number of targets. In this application, we require the constraint \eqref{eq:constr25} to hold for all targets $\mbf{x}_l(F+1)$ for some $\delta_i^l$, i.e $\det \Sigma_i^l \leq \delta_i^l$ where $\Sigma_i^l$ is the corresponding covariance matrix of target $l$. In what follows, we design mass functions that allow us to address large-scale estimation tasks
that involve large teams of robots and targets.
\vspace{-1.3 mm}

\subsection{Mass function $f_{\mcal{V}_i}$}
Let $L_{i,\max}^n$ denote the depth of tree $\mcal{G}_i$ at iteration $n$ and $\mcal{K}_{i,\max}^n$ be the set that collects indices $k$ that point to subsets $\mcal{V}_{i,k}$ for which there exists at least a node $\mbf{q}_i \in \mcal{V}_{i,k}$ that lays at depth $L_{i,\max}^n.$ Given the set $\mcal{K}_{i, \max}^n$ the mass function $f_{\mcal{V}_i}$ is designed as follows
\begin{equation}\label{eq:fknew}
f_{\mcal{V}_i}(k|\mcal{V}_i)=\left\{
                \begin{array}{ll}
                  p_{\mcal{V}_i}\frac{1}{|\mcal{K}_{i,\text{max}}^n|}, ~~~~~~~~~~~~\mbox{if}~k\in\mcal{K}_{i,\text{max}}^n\\
                  (1-p_{\mcal{V}_i})\frac{1}{|\mcal{V}_i\setminus\mcal{K}_{i,\text{max}}^n|},~\mbox{otherwise},
                \end{array}
              \right.
\end{equation}
where $p_{\mcal{V}_i} \in (0.5,1)$ is the probability of selecting any group $\mcal{V}_{i,k}$ indexed by $k \in \mcal{K}_{i,\max}^n$. 

\subsection{Mass function $f_{\mcal{U}_i}$}
The mass function $f_{\mcal{U}_i}$ is designed such that control inputs that drive robot $i$ with position $\mbf{p}_i(t)$ closer to target $j$ of predicted position $\hat{\mbf{x}}_{j}(t+1)$ at time $t+1$ are selected more often. If $\bm{u}_i^*$ is the control input that minimizes the geodesic distance between $\bm{p}_i(t+1)$ and $\hat{\bm{x}}_j(t+1)$, i.e $\bm{u}_i^* = \arg\min_{\bm{u} \in \mcal{U}_i}\left[d_{ij}= \parallel \bm{p}_i(t+1)-\hat{\bm{x}}_j(t+1) \parallel_g\right]$ the mass function $f_{\mcal{U}_i}$ is constructed as follows
\begin{equation}\label{eq:funew}
f_{\mcal{U}_i}(\bm{u}_i)=\left\{
                \begin{array}{ll}
                  p_{\mcal{U}_i}, ~~~~~~~~~~~~\mbox{if}~(\bm{u}_i=\bm{u}_i^*)\wedge(d_{ij} > R_i)\\
                  (1-p_{\mcal{U}_i})\frac{1}{|\mcal{U}_i|},~\mbox{otherwise},
                \end{array}
              \right.
\end{equation}
where (i) $p_{\mcal{U}_i} \in (0.5,1)$ is the probability that input $\mbf{u}_i$ is selected (ii) $\mbf{p}_i(t+1)$ is computed as per \eqref{eq:robot_dynamics} (iii) $\hat{\mbf{x}}_j(t+1)$ is the predicted position of target $j$ from the Kalman filter prediction step and (iv) $R_i$ denotes the sensing range. Once the predicted position $\hat{\mbf{x}}_j(t+1)$ is within the robot's sensing range, controllers are selected randomly.

\subsection{Mass function $f_{\mcal{S}_i}$}
The mass function $f_{\mcal{S}_i}$ is designed such that when more than one candidate nodes $\mbf{q}_j \in \mcal{Q}_{ij}$ are available, node $\mbf{q}_j^*$ with the smallest uncertainty, i.e $\vert \Sigma_j^* \vert = \arg\min_{\mbf{q}_j \in \mcal{Q}_{ij}} \vert \Sigma_j \vert$ is selected with higher probability. Given the set $\mcal{Q}_{ij}$ the mass function $f_{\mcal{S}_i}$ is the following

\begin{equation}\label{eq:fsnew}
f_{\mcal{S}_i}(\bm{q}_j)=\left\{
                \begin{array}{ll}
                  p_{\mcal{S}_i}, ~~~~~~~~~~~~\mbox{if}~\bm{q}_j=\bm{q}_j^*\\
                  (1-p_{\mcal{S}_i})\frac{1}{|\mcal{Q}_{ij}|},~\mbox{otherwise},
                \end{array}
              \right.
\end{equation}
where $p_{\mcal{S}_i} \in (0.5,1)$ is the probability that node $\mbf{q}_j$ is selected to provide covariance matrix $\Sigma_j(t)$.

\subsection{On-the-fly Target Assignment}
In Algorithm \ref{alg:target_assignment} we propose a target assignment process executed once the state $\qnew$ is created. Initially robot $i$ inherits the assigned target $l_{i,\text{rand}}$ of the parent node $\qrand$. Initially the sets $\mcal{T}_{i,\text{sorted}}, \mcal{T}_{i,\text{satisfied}}, \mcal{T}_{i,\text{unsatisfied}}$ and $\mcal{T}_{i,\text{occupied}}$ are created [Alg. \ref{alg:target_assignment}, lines \ref{trgt:sorted,satisfied}-\ref{trgt:occupied}]. %Only when the uncertainty condition $\det \Sigmanew^{l_{i,\text{rand}}} \leq \delta_i^{l_{i,\text{rand}}}$ is satisfied will robot $i$ place a request for a new target [Alg. \ref{alg:target_assignment}, line \ref{trgt:satisfied}], where $\Sigmanew^{l_{i,\text{rand}}}$ is the corresponding covariance matrix for target $l_{\text{i,rand}}$; otherwise it keeps the pre-assigned target $l_{i,\text{rand}}$ [Alg. \ref{alg:target_assignment}, line \ref{trgt:else}]. In the first case, the sets $\mcal{T}_{i,\text{sorted}}, \mcal{T}_{i,\text{satisfied}}, \mcal{T}_{i,\text{unsatisfied}}$ and $\mcal{T}_{i,\text{occupied}}$ are created [Alg. \ref{alg:target_assignment}, lines \ref{trgt:sorted,satisfied}-\ref{trgt:occupied}].
In particular, the set $\mcal{T}_{i,\text{sorted}}$ is the sorted set of indices $l \in \{1, \dots, M \}$ of increasing order according to the geodesic distance $\parallel \pnew-\hat{\mbf{x}}_l \parallel_g$. Moreover, the set $\mcal{T}_{i,\text{satisfied}}$ collects the indices of the satisfied targets, i.e $\mcal{T}_{i,\text{satisfied}} = \{l \mid \det \Sigmanew^{l} \leq \delta_i^l, \ l \in \{1,\dots, M\} \}$ [line \ref{trgt:sorted,satisfied}, Alg. \ref{alg:target_assignment}]. Given the aforementioned sets, we can compute the sorted set of the unsatisfied targets $\mcal{T}_{i,\text{unsatisfied}}$ [line \ref{trgt:unsatisfied}, Alg. \ref{alg:target_assignment}]. The set $\mcal{T}_{i,\text{occupied}}$ collects the indices $l_j$ that point to targets already assigned to neighbors $\mcal{N}_i$, i.e $\mcal{T}_{i,\text{occupied}} = \{l_j \mid \text{\textit{assigned target of }}\mbf{q}_j \in \mcal{Q}_{ij}, j \in \mcal{N}_i\}$ [line \ref{trgt:occupied}, Alg. \ref{alg:target_assignment}]. Once the sets are computed the robot sequentially seeks for the new target $l_{i,new}$ that is not already assigned to any of the neighbors and has not satisfied the uncertainty threshold [Alg. \ref{alg:target_assignment}, lines \ref{trgt:forloop}-\ref{trgt:same_target}].

\setlength{\textfloatsep}{0pt}
\begin{algorithm}[t]
\footnotesize
\caption{On-the-fly Target Assignment}
\LinesNumbered
\label{alg:target_assignment}
\KwIn{(i) $\delta_k, \forall k \in \{1, \dots M \}$, (ii) $\qnew$, (iii) current target $l_{\text{i,rand}}$}
\KwOut{assigned target $l_{i,\text{new}}$}
Compute sets $\mcal{T}_{i,\text{sorted}}$ and $\mcal{T}_{i, \text{satisfied}}$\; \label{trgt:sorted,satisfied}
Compute set $\mcal{T}_{i,\text{unsatisfied}} = \mcal{T}_{i, \text{sorted}} \backslash \mcal{T}_{i, \text{satisfied}}$\; \label{trgt:unsatisfied}
Compute set $\mcal{T}_{i, \text{occupied}}$\; \label{trgt:occupied}
$l_{i,\text{new}} = [ \ ]$\; \label{trgt:init_lnew} 
\For{$l \in \mcal{T}_{i,\text{unsatisfied}}$\label{trgt:forloop}}{
\If{$l \notin \mcal{T}_{i,\text{occupied}}$}{
$l_{i,\text{new}} = l$\;\label{trgt:if2}
break\;\label{trgt:break}
}
}
\If{$l_{i,\text{new}}$ is empty\label{trgt:if3}}{
$l_{i,\text{new}} = l_{i,\text{rand}}$\; \label{trgt:same_target}
}
\end{algorithm}

%\setlength{\textfloatsep}{1pt}
%\begin{algorithm}[t]
%\footnotesize
%\caption{On-the-fly Target Assignment}
%\LinesNumbered
%\label{alg:target_assignment}
%\KwIn{(i) $\delta_k, \forall k \in \{1, \dots M \}$, (ii) $\qnew$, (iii) current target $l_{\text{i,rand}}$}
%\KwOut{assigned target $l_{i,\text{new}}$}
%\If{$\det \Sigmanew^{l_{\text{i,rand}}} \leq \delta_i^l$ \label{trgt:satisfied}}{
%Compute sets $\mcal{T}_{i,\text{sorted}}$ and $\mcal{T}_{i, \text{satisfied}}$\; \label{trgt:sorted,satisfied}
%Compute set $\mcal{T}_{i,\text{unsatisfied}} = \mcal{T}_{i, \text{sorted}} \backslash \mcal{T}_{i, \text{satisfied}}$\; \label{trgt:unsatisfied}
%Compute set $\mcal{T}_{i, \text{occupied}}$\; \label{trgt:occupied}
%$l_{i,\text{new}} = [ \ ]$\; \label{trgt:init_lnew} 
%\For{$l \in \mcal{T}_{i,\text{unsatisfied}}$\label{trgt:forloop}}{
%\If{$l \notin \mcal{T}_{i,\text{occupied}}$}{
%$l_{i,\text{new}} = l$\;\label{trgt:if2}
%break\;\label{trgt:break}
%}
%}
%\If{$l_{i,\text{new}}$ is empty\label{trgt:if3}}{
%$l_{i,\text{new}} = l_{i,\text{rand}}$\; \label{trgt:same_target}
%}
%}
%\Else{
%$l_{i,\text{new}} = l_{i,\text{rand}}$\; \label{trgt:else}
%}
%\end{algorithm}
\section{Numerical Experiments}
In this section, we present numerical experiments that illustrate the performance of Algorithm \ref{alg:RRT} for the target localization and tracking problem, described in Section \ref{sec:biased}. We are interested in examining the scalability of Algorithm \ref{alg:RRT} and compare performances for different communication networks. All case studies have been implemented using Python 3.6 on a computer with Intel Core i10 1.3GHz and 16Gb RAM. 

In this section, each robot uses differential drive dynamics with $\mbf{p}_i(t)$ describing the position and orientation of robot $i$ at time $t$. The control commands use motion primitives $\{\nu_i, \omega_i\}, \ \nu_i \in \{0, 0.2, 1 \}\text{m/s}, \ \omega_i \in \{0, \pm 5, \pm 10, \pm 20, \pm 30, \\ \pm45, \pm 60 \}^\circ\text{/s}.$ %discretized with a sampling period $\tau$:
%\begin{align}
%\begin{bmatrix}
%    {p_{i,x}}_{t+1} \\
%    {p_{i,y}}_{t+1} \\
%    {\theta_i}_{t+1}
%    \end{bmatrix} = 
%    \begin{bmatrix}
%    {p_{i,x}}_{t} \\
%    {p_{i,y}}_{t} \\
%    {\theta_i}_{t}
%    \end{bmatrix} + 
%    \begin{bmatrix}
%    \nu \text{sinc}(\frac{\omega_i \tau}{2}) \cos({\theta_{i}}_t + \frac{\omega_i \tau}{2}) \\
%    \nu \text{sinc}(\frac{\omega_i \tau}{2}) \sin({\theta_{i}}_t + \frac{\omega_i \tau}{2}) \\
%    \tau \omega_i
%    \end{bmatrix}
%\end{align}
%where $\mathbf{p}_i(t) = [{p_{i,x}}_{t}, {p_{i,y}}_{t}, {\theta_i}_{t}]^T$ with ${p_{i,x}}_t, {p_{i,x}}_t, {\theta_i}_t$ denote the xy coordinates and the heading of robot $i$ at time t.
Furthermore, we assume that the robots are equipped with omnidirectional, range-only, line-of-sight sensors with limited range of 1m. The measurement received by robot $i$ regarding target $k$ is given by $y_{i,k} = l_{i,k}(t) + v_i(t)$ if $(l_{i,k}(t) \leq 1) \wedge (k \in \texttt{FOV}_i)$ where $l_{i,k}(t)$ is the distance between robot $i$ and target $k$, $\texttt{FOV}_i$ denotes the field-of-view of robot $i$ and $v_i(t) \sim \mcal{N}(0, \sigma^2(\mbf{p}_i(t), \mbf{x}_k(t))$ is the measurement noise with $\sigma = 0.25 l_{i,k}(t)$ if $l_{i,k}(t) \leq 1$; otherwise $\sigma$ is infinite. For separation principle to hold and offline control policies to be optimal we linearize the observation model about the predicted target position. In all case studies the targets are modeled as linear systems and the threshold parameter is selected to be $\delta_i^k=1.8\times 10^{-5}$ for all targets. The robots reside in the $10m \times 10m$ environment shown in Fig. \ref{fig:sim_trajectories}. The DKF parameters are chosen as $k_{ii} = 0.75$ and $k_{ij} = 0.25/\vert \mcal{N}_i \vert$ if $\det \Sigma_i^l \leq \det \Sigma_j^l$ for a target $l$; otherwise the values interchange.
\vspace{-1.3 mm}

\subsection{Scalability Performance}
In this Section we examine the scalability of the proposed distributed scheme with respect to the number of robots $N$ and targets $M$ for different communication networks. In Table \ref{table:time} we provide averaged results from ten Monte-Carlo trials for random and fully connected communication networks and the centralized approach in \cite{kantaros2019asymptotically}. Especially for the random communication networks we state the average degree (A.D) of the vertices pointing to the robots. The results come from a sequential implementation of Algorithm \ref{alg:RRT} and the time required by the "slowest" robot to build its tree is reported in Table \ref{table:time} for the distributed scheme. %reported numbers for the distributed method are  expected to be larger than the actually numbers if implemented in parallel. 

\begin{table}[t]
\caption{Scalability Analysis}
\label{table:time}
\centering
\begin{tabular}{l|l|l|l|l|}
\cline{2-5}
  & \multicolumn{2}{l|}{~ ~ ~ ~ \textbf{Connected}} 
  & \multicolumn{1}{l|}{\textbf{~ All-to-All}} 
  & \multicolumn{1}{l|}{\textbf{~ Centralized}} \\ \hline
\multicolumn{1}{|l|}{N/M}   & A.D & ~  Runtime / F & ~ Runtime / F & ~~ Runtime / F \\
\hline
\multicolumn{1}{|l|}{10/10} & 5.60 & 2.09 secs / 28   & 1.45 secs / 15 &  7.670 secs / 11 \\ \hline
\multicolumn{1}{|l|}{10/20} & 5.00 & 3.63 secs / 33   & 2.93 secs / 26 &  15.03 secs / 18        \\ \hline
\multicolumn{1}{|l|}{10/35} & 5.40 & 5.69 secs / 40   & 3.37 secs / 25 &  25.37 secs / 19    \\ \hline
\multicolumn{1}{|l|}{15/20} & 8.10 & 2.72 secs / 26   & 2.60 secs / 26 &  14.30 secs / 13        \\ \hline
\multicolumn{1}{|l|}{15/35} & 8.50 & 4.41 secs / 34   & 3.77 secs / 26 &  25.24 secs / 15  \\ \hline
\multicolumn{1}{|l|}{20/20} & 11.1 & 2.93 secs / 28   & 2.11 secs / 17 &  23.14 secs / 14     \\ \hline
\multicolumn{1}{|l|}{20/25} & 10.5 & 4.00 secs / 33   & 2.51 secs / 19 &  21.27 secs / 12        \\ \hline
\multicolumn{1}{|l|}{20/35} & 10.7 & 4.64 secs / 32   & 3.51 secs / 24 &  26.02 secs / 11     \\ \hline
\multicolumn{1}{|l|}{30/50} & 17.0 & 5.32 secs / 26   & 4.29 secs / 19 &  46.02 secs / 10 \\ \hline
\end{tabular}
%\vspace{-5.5 mm}
%\vspace{-2 mm}
\end{table}

%%%% total time table %%%%%
%\begin{table}[t]
%\caption{Scalability Analysis}
%\label{table:time}
%\centering
%\begin{tabular}{l|l|l|l|l|}
%\cline{2-5}
%  & \multicolumn{2}{l|}{~ ~ ~ ~ \textbf{Connected}} 
%  & \multicolumn{1}{l|}{\textbf{~ All-to-All}} 
%  & \multicolumn{1}{l|}{\textbf{~ Centralized}} \\ \hline
%\multicolumn{1}{|l|}{N/M}   & A.D & ~  Runtime / F & ~ Runtime / F & ~~ Runtime / F \\
%\hline
%\multicolumn{1}{|l|}{10/10} & 5.60 & 16.78 secs / 26  & 11.29 secs / 21 &  8.171 secs / 12 \\
%\hline
%\multicolumn{1}{|l|}{10/20} & 5.00 & 24.83 secs / 33   & 13.22 secs / 19 &  14.88 secs / 17        \\ \hline
%\multicolumn{1}{|l|}{10/35} & 5.40 & 35.97 secs / 35   & 21.27 secs / 24  &  26.08 secs / 21    \\ \hline
%\multicolumn{1}{|l|}{15/20} & 8.10 & 34.91 secs / 31   & 25.75 secs / 25   &  17.40 secs / 15        \\ \hline
%\multicolumn{1}{|l|}{15/35} & 8.50 & 47.10 secs / 34   & 36.94 secs / 29   &  25.04 secs / 16  \\ \hline
%\multicolumn{1}{|l|}{20/20} & 11.1 & 37.30 secs / 26   &  22.66 secs / 15    & 21.03 secs / 13     \\ \hline
%\multicolumn{1}{|l|}{20/25} & 10.5 & 51.93 secs / 35   & 35.42 secs / 24   & 22.16 secs / 13        \\ \hline
%\multicolumn{1}{|l|}{20/35} & 10.7 & 1.007 mins / 31   & 45.82 secs / 25   & 25.47 secs / 12     \\ \hline
%\multicolumn{1}{|l|}{30/50} & 17.0 & 1.73 mins / 28  & 1.48 mins / 24  & 50.08 secs / 11 \\ \hline
%\end{tabular}
%\vspace{-5.5 mm}
%\vspace{-2 mm}
%\end{table}

\begin{figure}[t]
    \centering
    \includegraphics[width=0.3\textwidth]{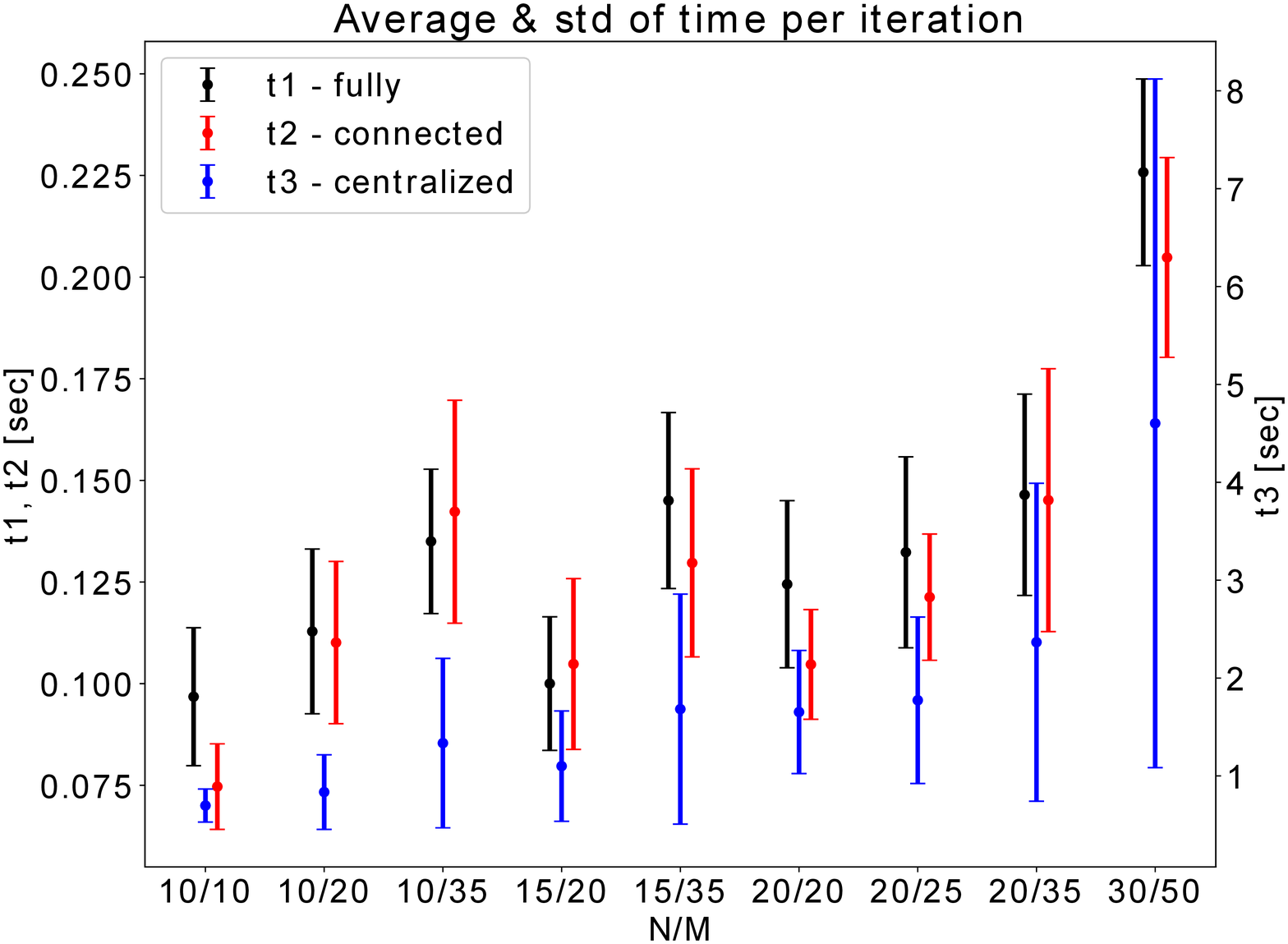}
    \setlength{\belowcaptionskip}{-10pt}
    \caption{This Figure illustrates the average time needed for a completion of one iteration across the entire planning horizon $F$ for $N$ robots, $M$ targets for three cases (i) random connected network (red), (ii) fully connected network (black) and (iii) centralized method (blue).}
    \label{fig:avg_time_comparison}
\end{figure}

We can notice that feasible paths are computed very fast for all connectivity cases regardless of the number of robots and targets. The planning horizon $F$ of the random connected networks is always larger since any transmitted information needs more timesteps until diffused. The centralized approach returns the smallest $F$, as in contrast to DKF the update rule of Kalman Filter fuses measurement models of all robots and therefore the desired uncertainty threshold is reached faster. In Fig. \ref{fig:avg_time_comparison} the average time of completion of one iteration is presented for all scenarios of Table \ref{table:time}. We can see that the distributed approach improves the computational complexity per iteration by at least a factor of 10. %The computational complexity of the centralized approach is much higher than that of Algorithm \ref{alg:RRT} since in a single iteration $N$ new states $\mbf{p}_{\text{new}}$ are created and a Kalman filter update requires $N$ measurement models.

\subsection{Different communication networks}
\begin{figure}[t]
     \centering
     \begin{subfigure}[b]{0.23\textwidth}
         \centering
         \includegraphics[width=0.9\textwidth]{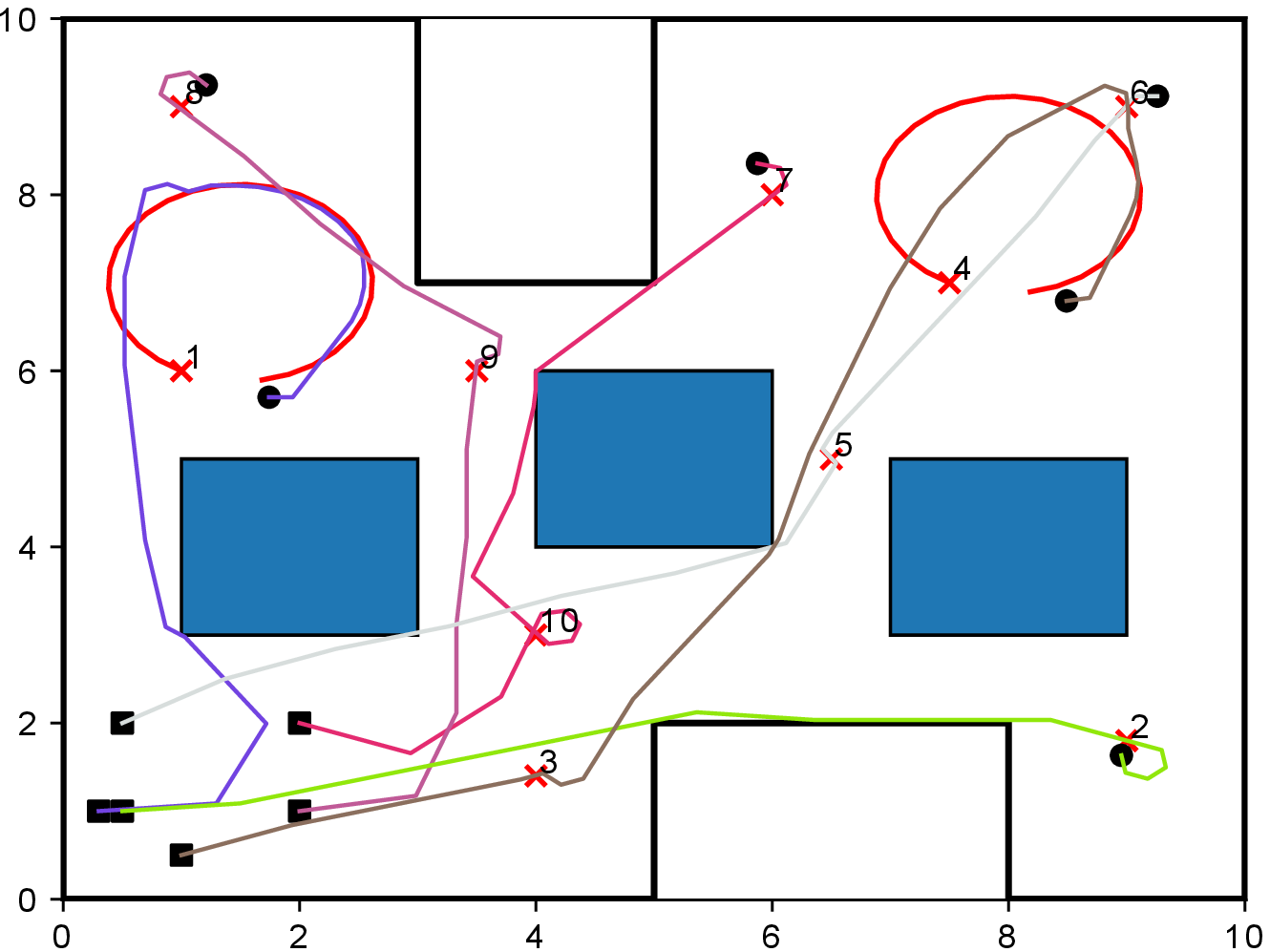}
         \caption{}
         \label{fig:sim_fully_connected}
     \end{subfigure}
     \hfill
    \begin{subfigure}[b]{0.23\textwidth}
         \centering
         \includegraphics[width=0.9\textwidth]{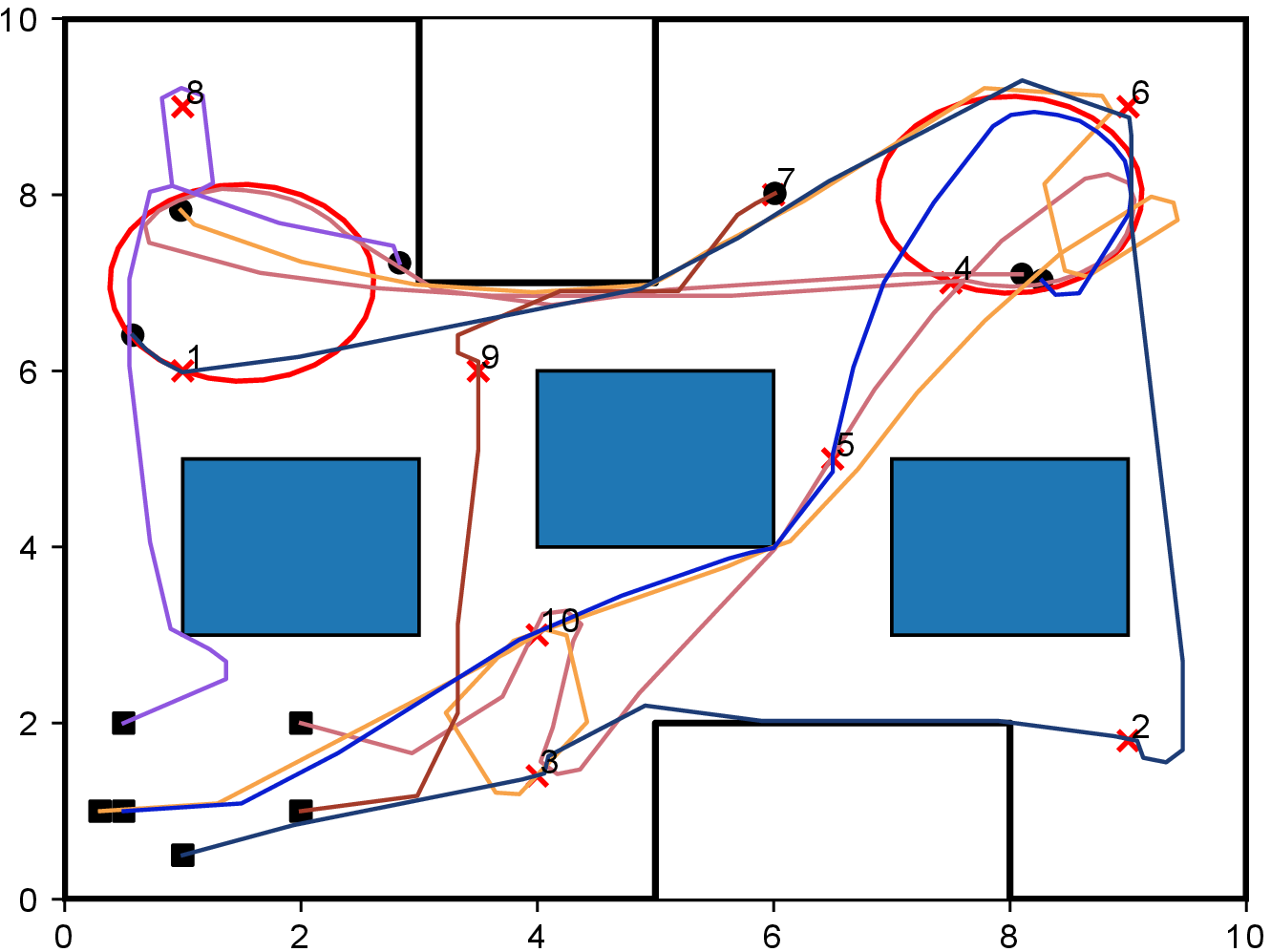}
         \caption{}
         \label{fig:sim_connected}
     \end{subfigure}
     \vfill
     \begin{subfigure}[b]{0.23\textwidth}
         \centering
         \includegraphics[width=0.9\textwidth]{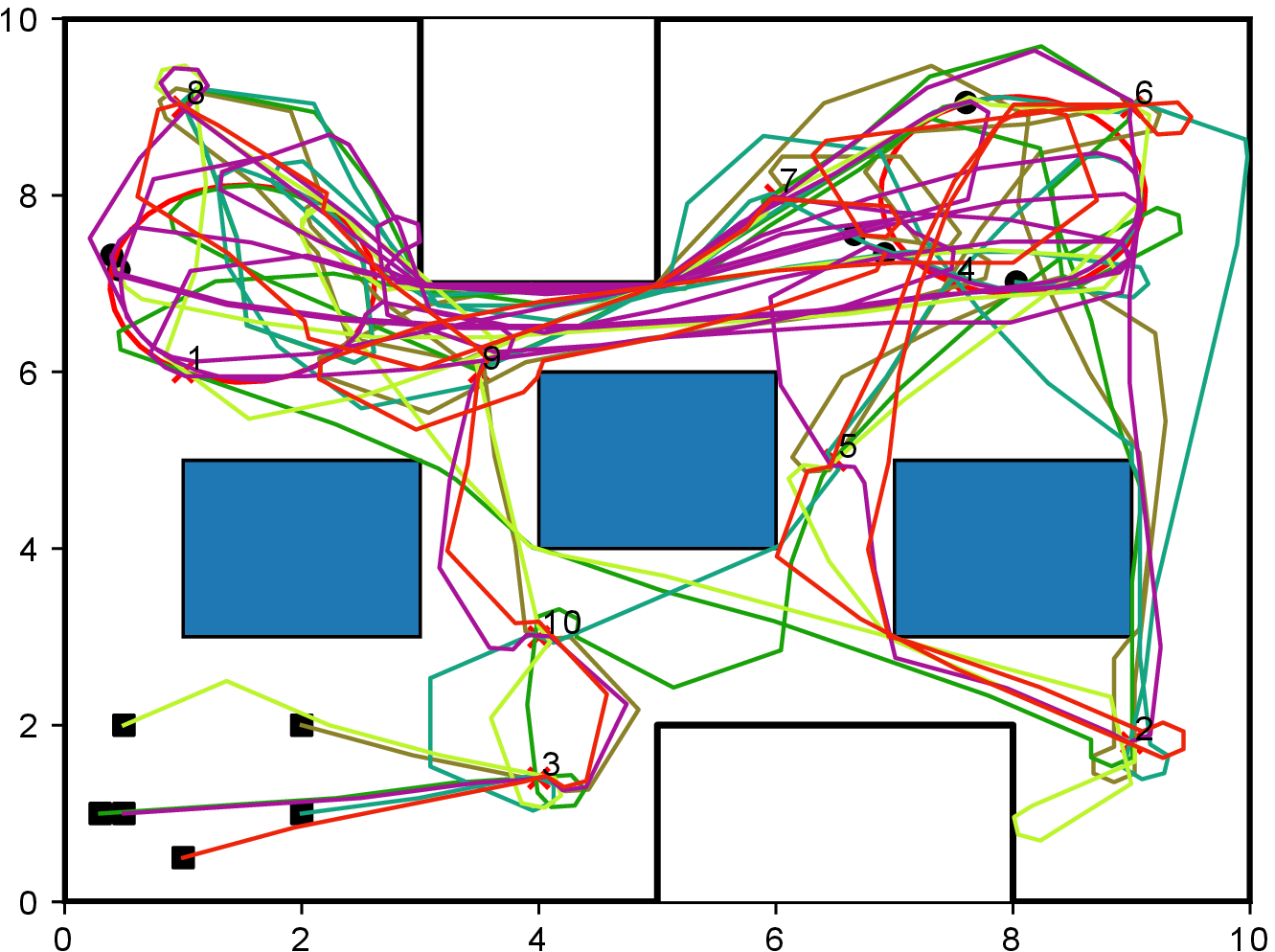}
         \caption{}
         \label{fig:sim_disconected}
     \end{subfigure}          \begin{subfigure}[b]{0.23\textwidth}
         \centering
         \includegraphics[width=0.9\textwidth]{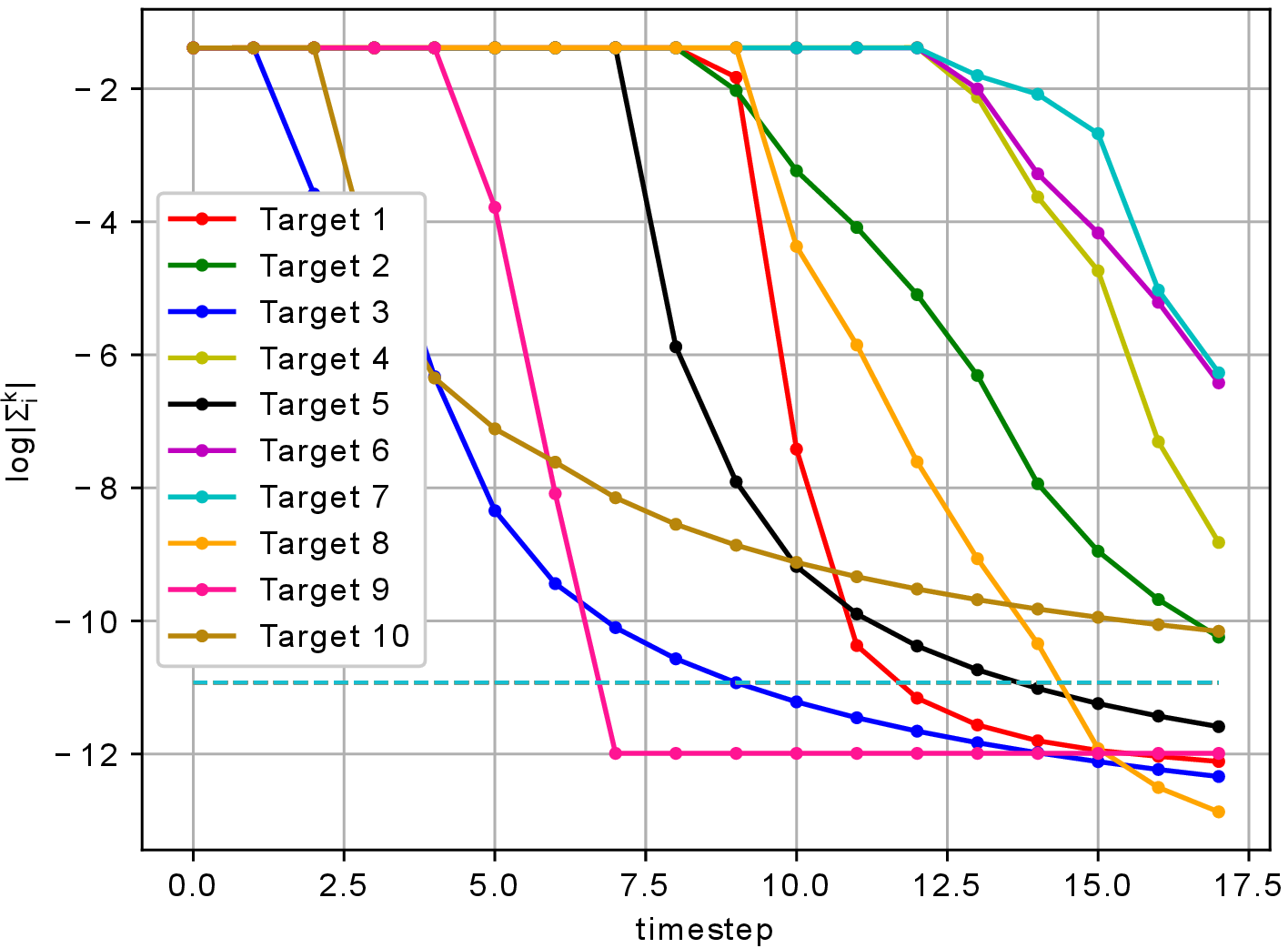}
         \caption{}
         \label{fig:sim_uncertainty}
     \end{subfigure}
     \setlength{\belowcaptionskip}{-18pt}
     \caption{Trajectories of $N=6$ robots and $M=10$ targets for three communication networks (a) fully connected, (b) connected and (c) no communication. The initial positions of the targets are depicted as red crosses and the trajectory of moving targets can be seen in solid red color. The initial and final position of the robots are depicted by black squares and circles respectively. In Fig. \ref{fig:sim_uncertainty} the uncertainties $\det \Sigma_i^k$ are presented for the "purple" robot of Fig. \ref{fig:sim_fully_connected}. The dashed line depicts the threshold $\delta_i^k$.}
     \label{fig:sim_trajectories}
\end{figure}

In this section we examine the performance of Algorithm \ref{alg:RRT} for different communication graph densities. Specifically, in Fig. \ref{fig:sim_trajectories}, the trajectories for three different communication scenarios are presented for $N=6$ robots and $M=10$ targets (i) fully connected graph (ii) connected graph with three edges average degree and (iii) no communication. In the current example, the planning terminates once a robot realizes that the targets have been identified either by itself and/or by any of the neighboring robots.

%\begin{figure}[h]
%    \centering
%    \includegraphics[width = 0.3\textwidth]{figures/simulations/time_per_iter.eps}
%    \caption{Caption}
%    \label{fig:my_label}
%\end{figure}

 The density of the communication network affects the returned paths and the planning horizon $F$. Specifically, in the case of a fully connected network as in Fig. \ref{fig:sim_fully_connected} % the covariance matrix $\Sigma_i(t)$ of any robot $i$ is received after one timestep by all participating robots. The speed of fusing beliefs in this case is high and thus robots reduce their uncertainty even for targets that have not been visited by themselves. The same idea extends in the case of a random connected network as in Fig.
 the speed of fusing beliefs is high and thus robots reduce their uncertainty even for targets that have not been visited by themselves. The same idea extends in the case of a random connected network as in Fig. \ref{fig:sim_connected} with the only difference being that the fusion of beliefs has a smaller rate and therefore robots may revisit targets that have already been satisfied. In fact the less dense a graph is the more the planning horizon $F$ increases. In the case of no communication in Fig. \ref{fig:sim_disconected} each robot plans completely on its own and therefore every target is visited. Kalman Filter updates may lead to an increase of the uncertainty over a dynamic target even if in the past it was well estimated. The robots then as in Fig. \ref{fig:sim_disconected} oscillate between the two moving targets leading to a further increase of their planning horizon. The returned planning horizons of the aforementioned cases are $F = [24, 29, 120]$, where for the disconnected network the maximum planning horizon out of the individual planning horizons of each robot is stated.
 
In Fig. \ref{fig:sim_uncertainty} the evolution of the uncertainties $\det \Sigma_i^k$ over time is depicted for the "purple" robot of the fully connected case of Fig. \ref{fig:sim_fully_connected}. Thanks to the connectivity capabilities, we can see that the current robot has not reduced the uncertainty for the entire group of targets since it realizes that there exist other robots that took the responsibility of identifying the remaining targets.

%\begin{figure}
%    \centering
%    \includegraphics[width=0.4\textwidth]{figures/simulations/large_trajectories_trial_0.eps}
%    \caption{Caption}
%    \label{fig:my_label}
%\end{figure}

\section{Conclusions}
A new distributed sampling-based algorithm for active information gathering is introduced in this paper followed by formal guarantees. The proposed scheme is tested via extensive simulations showing that feasible paths are quickly computed for large-scale problems, improving significantly also the computational time of existing scalable centralized method. In the future, time-varying communication graphs will be incorporated in the proposed framework.

\bibliographystyle{IEEEtran}
\bibliography{bibliography}

\section{Appendix: Proof of Separation Principle,  Completeness \& Optimality}\label{sec:appendix}

To prove Theorems \ref{thm:probCompl}, \ref{thm:asOpt} we need to prove the following results.

\begin{lemma}(\textbf{Sampling} $\mcal{V}^n_{i, k_{\text{rand}}}$)
Consider any subset $\mcal{V}_{i,k}^n$, any fixed iteration index $n$ and any fixed $k\in\{1,\dots,K_{i,n}\}$. Then, there exists an infinite number of subsequent iterations $n+w$,  where $w\in\mathcal{W}$ and $\mathcal{W}\subseteq\mathbb{N}$ is a subsequence of $\mathbb{N}$, at which the subset $\mcal{V}_{i,k}^n$ is selected to be the set $\mcal{V}_{i,k_{\text{rand}}}^{n+w}$. 
\label{lem:qrand}
\end{lemma}

\begin{lemma}[\textbf{Sampling} $\bm{u}_{i,\text{new}}$]
Consider any  subset $\mcal{V}_{i, k_{\text{rand}}}^n$ selected by $f_{\mcal{V}_i}$ and any fixed iteration index $n$. Then, for any given control input $\bm{u}\in\mcal{U}_i$, there exists an infinite number of subsequent iterations $n+w$, where $w\in\mathcal{W}'$ and $\mathcal{W}'\subseteq\mathcal{W}$ is a subsequence of the sequence of $\mathcal{W}$ defined in Lemma \ref{lem:qrand}, at which the control input $\bm{u}\in\mcal{U}_i$ is selected to be $\bm{u}_{i,\text{new}}^{n+w}$.
\label{lem:qnew}
\end{lemma}

\begin{proof}
The proofs of Lemmas \ref{lem:qrand}, \ref{lem:qnew} resemble the proof of \cite[Appendix A]{kantaros2019asymptotically} and are omitted.
\end{proof}

\begin{lemma}(\textbf{Sampling} $\mbf{q}_j$)
Consider any subset $\mcal{V}_{i, k_{\text{rand}}}^n$, any control input $\bm{u}_{i, \text{new}}$ and any fixed iteration $n$. Then for any $\mbf{q}_j \in \mcal{Q}_{ij}^n$ there exists an infinite number of subsequent iterations $n+w$, where $w \in \mcal{W}''$ and $\mcal{W}'' \subseteq \mcal{W}'$ is a subsequence of the sequence $\mcal{W}'$ defined in Lemma \ref{lem:qnew}, at which node $\mbf{q}_j$ is selected to be added in the set $\mcal{S}_{i,new}$.
\label{lem:Snew}
\end{lemma}

\begin{proof}
%This proof resembles the proof of Lemma \ref{lem:qrand} and is omitted.
%
Define the infinite sequence of events $A^{\text{new}}=\{A^{\text{new},n+w}(\bm{q}_j)\}_{w=0}^{\infty}$, for $\bm{q}_j\in\mcal{Q}_{ij}^n$, where $A^{\text{new},n+w}(\bm{q}_j)=\{\bm{q}_j^{n+w}=\bm{q}_j\}$, for $w\in\mathbb{N}$, denotes the event that at iteration $n+w$ of Algorithm \ref{alg:RRT} node $\bm{q}_j\in\mcal{Q}_{ij}^n$ is selected by the sampling function to be the node $\bm{q}_{j}^{n+w}$, given the subset $\mcal{V}_{i, k_{\text{rand}}}^n\in\mcal{V}_{k_{\text{rand}}}^{n+w}$ and control input $\mbf{u}_{i,\text{new}}$. Moreover, let $\mathbb{P}(A^{\text{new},n+w}(\bm{q}_j))$ denote the probability of this event, i.e., $\mathbb{P}(A^{\text{new},n+w}(\bm{q}_j))=f_{\mcal{Q}_i}^{n+w}(\bm{q}_j| \mcal{Q}_{ij}^{n+w})$. 
Now, consider those iterations $n+w$ with $w\in\mathcal{W'}$ such that $k_{\text{rand}}^{n+w}=k_{\text{rand}}^n$ and $\mbf{u}_{i,\text{new}}^{n+w} = \mbf{u}_{i,\text{new}}^n$ by Lemma \ref{lem:qnew}. We will show that the series $\sum_{w\in\mcal{W'}}\mathbb{P}(A^{\text{new},n+w}(\bm{q}_j))$ diverges and then we will use Borel-Cantelli lemma  to show that any given $\bm{q}_j\in\mcal{U}$ will be selected infinitely often to be $\bm{q}_j^{n+w}$.
By Assumption \ref{as:fq}(i)  we have that $\mathbb{P}(A^{\text{new},n+w}(\bm{q}_j))=f_{\mcal{Q}_i}^{n+w}(\bm{q}_j | \mcal{Q}_{ij}^{n+w})$ is bounded below by a strictly positive constant $\xi>0$ for all $w\in\mcal{W}'$. 
Therefore, we have that $\sum_{w\in\mcal{W}'}\mathbb{P}(A^{\text{new},n+w}(\bm{q}_j))$ diverges, since it is an infinite sum of a strictly positive constant term. Using this result along with the fact that the events $A^{\text{new},n+w}(\bm{q}_j)$ are independent, by Assumption \ref{as:fq} (ii), we get that $\mathbb{P}(\limsup_{w\to\infty} A^{\text{new},n+w}(\bm{q}_j))=1,$ due to the Borel-Cantelli lemma. In words, this means that the events $A^{\text{new},n+w}(\bm{q}_j)$ for $w\in\mathcal{W}'$ occur infinitely often. Thus, given any subset $\mcal{V}_{i,k_{\text{rand}}}^n$, any control input $\bm{u}_{i, \text{new}}^n$ for all $q_j \in \mcal{Q}_{ij}^n$ and for all $n\in\mathbb{N}_{+}$, there exists an infinite subsequence $\mathcal{W}'' \subseteq \mathcal{W}'$ so that for all $w\in\mathcal{W}''$ it holds $\bm{q}_j^{n+w}=\bm{q}_j$, completing the proof.
\end{proof}

Before stating the next result, we first define the \textit{reachable} state-space of a state $\bm{q}_i(t)=[\bm{p}_i(t),\Sigma_i(t), \mcal{S}_i(t)]\in\mcal{V}_{i,k}^n$, denoted by $\mcal{R}(\bm{q}_i(t))$ that collects all states $\bm{q}_i(t+1)=[\bm{p}_i(t+1), \Sigma_i(t+1), \mcal{S}_i(t+1)]$ that can be reached within one time step from $\bm{q}_i(t)$.

\begin{cor}[Reachable set $\mcal{R}(\bm{q}_i(t))$]\label{cor:reach}
Given any state $\bm{q}_i(t)=[\bm{p}_i(t),\Sigma_i(t),\mcal{S}_i(t)]\in\mcal{V}_{i,k}^n$, for any $k\in\{1,\dots,K_{i,n}\}$, Algorithm \ref{alg:RRT} will add to $\mcal{V}_i^n$ all states that belong to the reachable set $\mcal{R}(\bm{q}_i(t))$ will be added to $\mcal{V}_i^{n+w}$, with probability 1, as $w\to\infty$, i.e., $\lim_{w\rightarrow\infty} \mathbb{P}\left(\{\mcal{R}(\bm{q}_i(t))\subseteq\mathcal{V}_i^{n+w}\}\right)=1.$
Also, edges from $\bm{q}_i(t)$ to all reachable states $\bm{q}_i'(t+1)\in\mcal{R}(\bm{q}_i(t))$ will  be added to $\mcal{E}_i^{n+w}$, with probability 1, as $w\to\infty$, i.e., $\lim_{w\rightarrow\infty} \mathbb{P}\left(\{\cup_{\bm{q}_i'\in\mcal{R}(\bm{q}_i)}(\bm{q}_i,\bm{q}_i')\subseteq\mathcal{E}_i^{n+w}\}\right)=1.$
\end{cor}
\begin{proof}
The proof straightforwardly follows from Lemmas \ref{lem:qrand}-\ref{lem:Snew} and is omitted. 
\end{proof}

{\bf{Proof of Theorem \ref{thm:probCompl}}:}  
By construction of the path $\bm{q}_{i,0:F}$, it holds that $\bm{q}_i(f)\in\mcal{R}(\bm{q}_i(f-1))$, for all $f\in\{1,\dots,F\}$. Since $\bm{q}_i(0)\in\mcal{V}_i^1$, it holds that all states $\bm{q}_i\in\mcal{R}(\bm{q}_i(0))$, including the state $\bm{q}_i(1)$, will be added to $\mcal{V}_i^n$ with probability $1$, as $n\to\infty$, due to Corollary \ref{cor:reach}.
Once this happens, the edge $(\bm{q}_i(0),\bm{q}_i(1))$ will be added to set of edges $\mcal{E}_i^n$ due to Corollary \ref{cor:reach}.
Applying Corollary \ref{cor:reach} inductively, we get that $\lim_{n\rightarrow\infty} \mathbb{P}\left(\{\bm{q}_i(f)\in\mathcal{V}_i^{n}\}\right)=1$ and $\lim_{n\rightarrow\infty} \mathbb{P}\left(\{(\bm{q}_i(f-1), \bm{q}_i(f))\in\mathcal{E}_i^{n}\}\right)=1$, for all $f\in\{1,\dots,F\}$ meaning that the path $\bm{q}_{i,0:F}$ will be added to the tree $\mcal{G}_i^n$ with probability $1$ as $n\to\infty$ completing the proof.

{\bf{Proof of Theorem \ref{thm:asOpt}}:}
The proof of this result straightforwardly follows from Theorem \ref{thm:probCompl}. Specifically, recall from
Theorem \ref{thm:probCompl} that Algorithm \ref{alg:RRT} can find any feasible path and,
therefore, the optimal path as well, with probability 1, as $n \to \infty$, completing the proof.

{\bf{Proof of Theorem \ref{thm:time}}:}
For the proof, a fixed iteration is considered and the computational complexity is analyzed with parameters the number of robots $N$ and maximum degree of the vertices $d_{\max}$ of the communication graph $G$, where $d_{\max} \leq N-1$. Let's assume that both methods need to expand $k$ nodes $\mbf{q}_{\text{rand}}$ and robot $i$ is the one with the most neighbors, i.e $d_{\max} = \vert \mcal{N}_i \vert$. %Given nodes $\mbf{q}_{\text{rand}}$, 
In the centralized approach \cite{kantaros2019asymptotically} each state $\mbf{p}_{\text{rand}}$ describes $N$ concatenated states $\mbf{p}_{i, \text{rand}}$. The computation then of $\mbf{p}_{\text{new}}$ is of order $\mcal{O}(kN)$, while for Algorithm \ref{alg:RRT} is $\mcal{O}(k)$. The Kalman filter update rule in \cite{kantaros2019asymptotically} is computed based on $N$ measurement models, while in Algorithm \ref{alg:RRT} each state $\pnew$ uses information from $\vert \mcal{N}_i \vert$ neighbors. The respective Kalman Filter complexities then are $\mcal{O}(N)$ and $\mcal{O}(k \vert \mcal{N}_i \vert)$. Note that sampling a single-query from a discrete distribution and checking membership of $\mbf{q}_{\text{new}}$ in a subset $\mcal{V}_k$ depends on parameters which are not part of our analysis and are not taken into account. Summing-up, the computational complexity per iteration of centralized approach \cite{kantaros2019asymptotically} and Algorithm \ref{alg:RRT} are $\mcal{O}(kN + N) = \mcal{O}(kN)$ and $ \mcal{O}(k + k d_{\text{max}}) = \mcal{O}(kd_{\text{max}})$ respectively, completing the proof.
%Each $\qrand$ expects from neighbor $j$ the set of admissible nodes $\mcal{Q}_{ij}$. The construction of the set $\mcal{Q}_{ij}$ requires $\mcal{O}(1)$ time when the nodes $\mbf{q}_j$ are described by hash-tables and update their children. Therefore, the worst-case time complexity of the distributed approach per iteration is $ \mcal{O}( 2^n \vert \mcal{N}_i \vert)$. In the centralized approach, $\mbf{q}_{\text{rand}}$ contains in total $N$ nodes one for each robot and the complexity per iteration is $\mcal{O}(2^n N)$. Comparing the two approaches the proposed distributed approach is $d_{\max}/N$ times faster than the centralized per iteration, completing the proof.  
\end{document}